%% file: main.tex
\documentclass[10pt,twocolumn,letterpaper]{article}
\usepackage{hhline}
\usepackage{lscape}
\usepackage{array}
\usepackage{ifthen}
\usepackage{colortbl}
\usepackage{multirow}
\usepackage{siunitx}
\usepackage{xspace}
\usepackage{algorithm}
\usepackage{algpseudocode}
\usepackage{graphicx} 
\usepackage{dsfont}
\usepackage{caption} 
\usepackage{wrapfig}
\usepackage{xcolor}
\usepackage{pifont} 
\usepackage{algpseudocode}
\usepackage{float}
\usepackage{graphicx}
\usepackage{booktabs}
\usepackage{placeins}
\usepackage[runningheads]{cvpr} 
\renewcommand{\cvprrunningauthor}{Ray et al.}
\renewcommand{\cvprrunningtitle}{\BEACON}

\input{preamble}

\definecolor{cvprblue}{rgb}{0.21,0.49,0.74}
\usepackage[pagebackref,breaklinks,colorlinks,allcolors=cvprblue]
{hyperref}

\def\confName{CVPR}
\def\confYear{2027}

\title{\BEACON: Learning Observation-Conditioned Starting Points\\for Guided Conditional Flow Matching in Microscopy}

\author{
\textbf{Anirban Ray}$^{1,2}$ \quad
\textbf{Florian Jug}$^{1}$ \\
$^{1}$Human Technopole, Italy \quad
$^{2}$Technische Universität Dresden, Germany \\
{\tt\small \{anirban.ray, florian.jug\}@fht.org}
}

\input{figures}

\begin{document}
\raggedbottom
\maketitle
\input{sec/0_abstract}
\input{sec/1_intro}

\input{sec/2_RelatedWork}
\input{sec/3_Method}
\input{sec/4_Experiments}
\input{sec/5_Results}
\input{sec/6_Discussion}
\input{sec/9_llm_statement} 
{
    \small
    \bibliographystyle{ieeenat_fullname}
    \bibliography{main}
}

\input{sec/8_Appendix}

\end{document}

%% file: preamble.tex
\newcommand{\HazeMatching}{\mbox{\textsc{HazeMatching}}\xspace}

\newcommand{\ResMatching}{\mbox{\textsc{ResMatching}}\xspace}
\newcommand{\BEACON}{\mbox{\textsc{Beacon}}\xspace}
\newcommand{\miniheadline}[1]{\par\smallskip\noindent\textbf{#1.}\hspace{0.5em}}

\newcommand{\HVAE}{\mbox{\textsc{LVAE}}\xspace}

\newcommand{\UNet}{\mbox{\textsc{U-Net}}\xspace}

%% file: figures.tex
\newcommand{\figTeaserMain}{
\begin{figure}[t]
  \centering
    \includegraphics[width=1.0\linewidth]{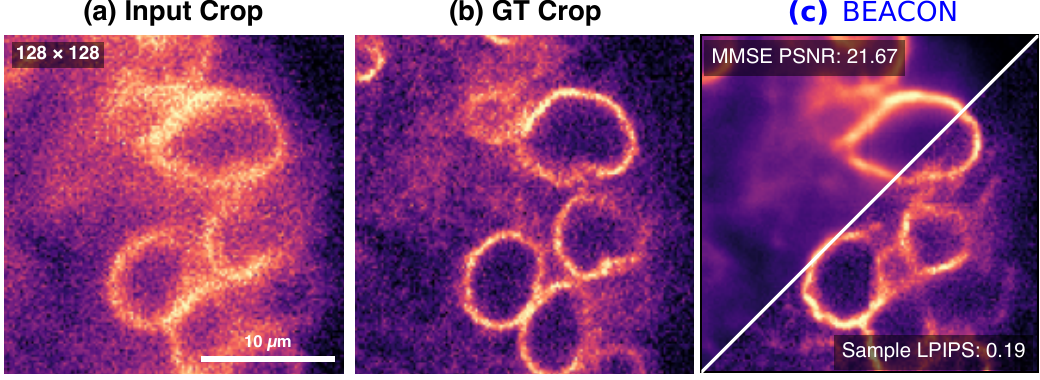} 
\caption{\textbf{\BEACON balances reconstruction fidelity and perceptual realism.}
(a,b) Degraded input and corresponding ground truth.
(c) \BEACON ($K=16$) reconstruction showing the MMSE estimate (upper-left) and a posterior sample (lower-right).}
  \label{fig:teaser_main}
\end{figure}
}

\newcommand{\figTraining}{
\begin{figure*}[h]
  \centering
    \includegraphics[width=0.8\textwidth]{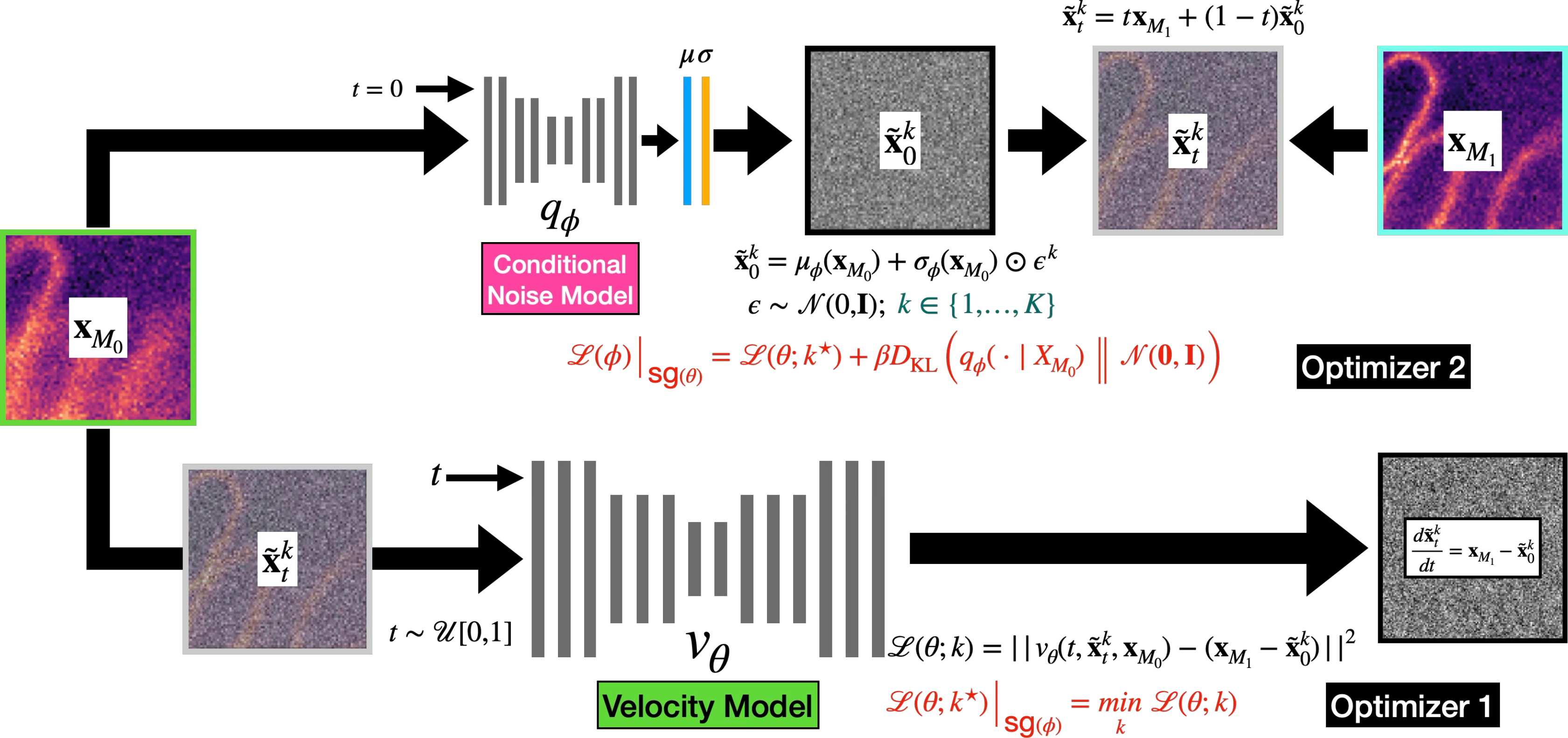}
    \caption{\textbf{Overview of the \BEACON training framework.}
Given a degraded microscopy observation $\mathbf{x}_{M_0}$, the conditional
noise model $q_\phi$ predicts spatially varying Gaussian parameters
$(\mu_\phi,\sigma_\phi)$, from which $K$ candidate latents
$\tilde{\mathbf{x}}_0^k$ are sampled.
Each candidate is linearly interpolated with the corresponding
clean target $\mathbf{x}_{M_1}$ to construct
$\tilde{\mathbf{x}}_t^k = t\mathbf{x}_{M_1} + (1-t)\tilde{\mathbf{x}}_0^k$,
where $t \sim \mathcal{U}[0,1]$.
The conditional flow model $v_\theta$ learns the associated
velocity field by minimizing the flow-matching loss
$\mathcal{L}(\theta;k)$.
A best-of-$K$ selection identifies the candidate $k^*$ with
the minimum flow-matching loss, which is used to update the
flow model through \textbf{Optimizer 1}.
The conditional noise model is updated separately through
\textbf{Optimizer 2}, using the selected flow-matching loss
with gradients stopped through $\theta$, together with a
KL regularization term that constrains the learned conditional
noise distribution toward the standard Gaussian prior.
This alternating optimization encourages adaptive
noise--target couplings while preserving the stochasticity
of the latent distribution.}
  \label{fig:overview}
\end{figure*}
}

\newsavebox{\nipdbox}
\newcommand{\nipdlines}{10}
\newcommand{\ranking}{
\begin{figure}[t]
\sbox{\nipdbox}{\includegraphics[width=0.42\linewidth]{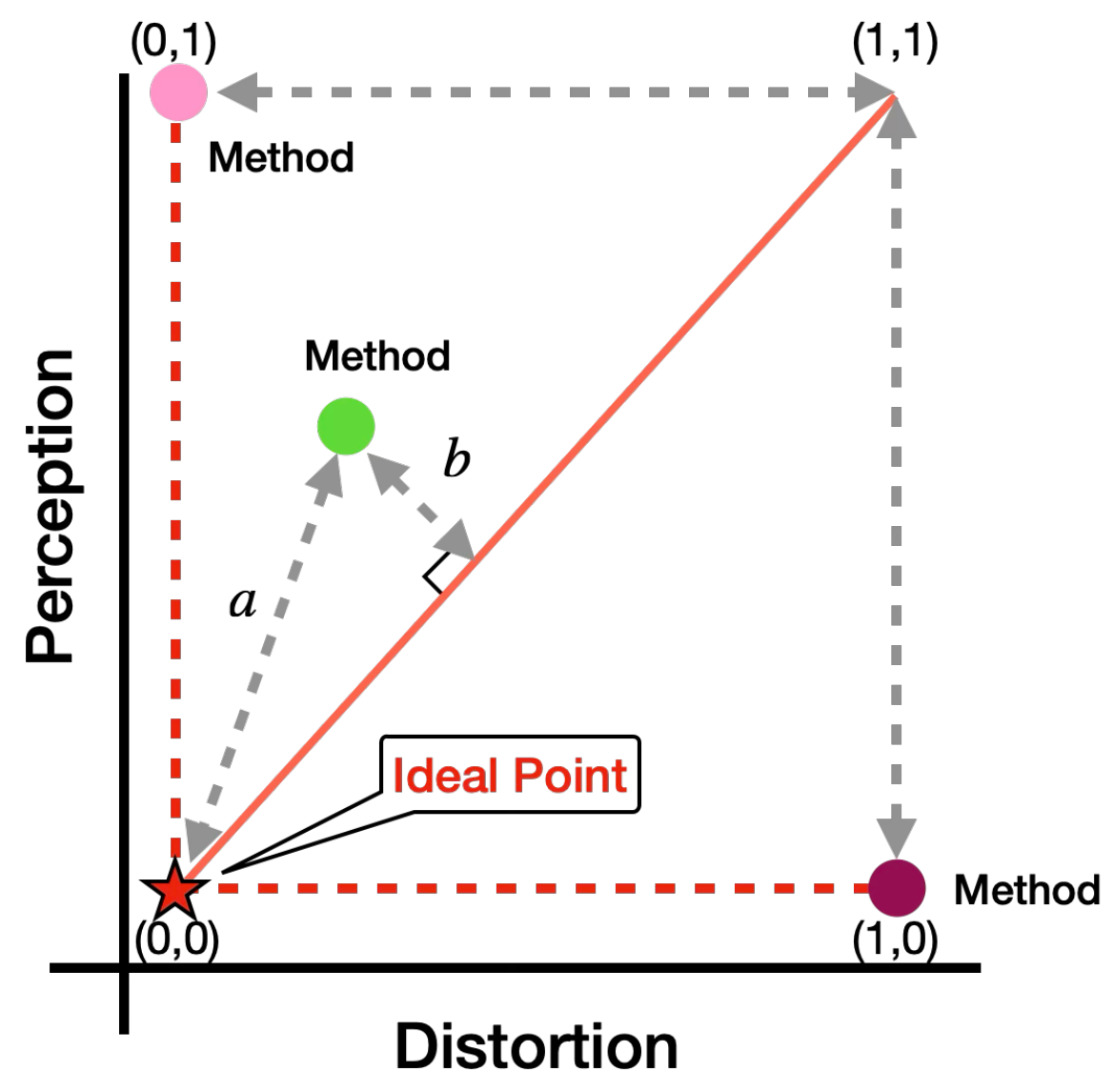}}%
\noindent\raisebox{-\height}[0pt][0pt]{\usebox{\nipdbox}}\par\vspace{-\baselineskip}%
\captionsetup{skip=0pt}%
\caption[Illustration of the Normalized Ideal-Point Distance (NIPD) ranking metric.]{\hangindent=\dimexpr\wd\nipdbox+0.8em\relax\hangafter=-\nipdlines\relax\textbf{Illustration of the Normalized Ideal-Point
Distance (NIPD) ranking metric.}
For each perception--distortion plot, the ideal point is
defined by the best distortion and perceptual scores
achieved independently among all compared methods.
Both metrics are normalized to $[0,1]$, placing the
ideal point at $(0,0)$. For each method, NIPD is
computed as $a+b$, where $a$ is the Euclidean distance
to the ideal point and $b$ is the perpendicular distance
to the diagonal representing equal normalized distortion
and perceptual error. Lower NIPD scores indicate a
more favorable perception--distortion trade-off.}
  \label{fig:nipd}
\end{figure}
}

\newcommand{\figRankMain}{
\begin{figure*}[t]
  \centering
  \includegraphics[width=0.9\textwidth]{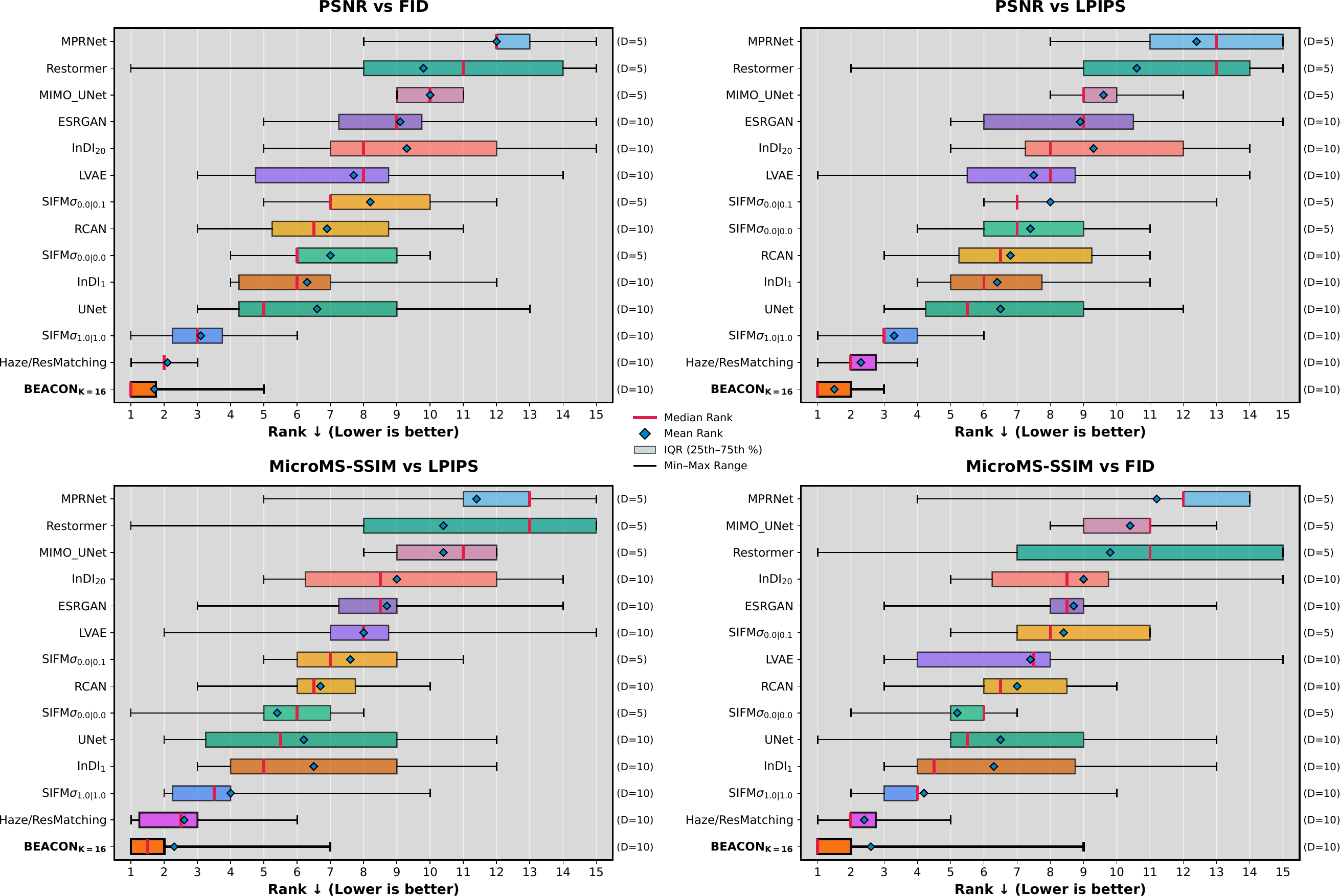}
\caption{\textbf{Aggregate perception--distortion rankings across microscopy restoration datasets.}
Distributions of Normalized Ideal-Point Distance (NIPD; Eq.~\eqref{eq:pd_rank_score}) ranks across datasets for four metric pairs: PSNR--FID (top-left), PSNR--LPIPS (top-right), MicroMS-SSIM--LPIPS (bottom-left), and MicroMS-SSIM--FID (bottom-right).
Methods are ordered by median rank (lower is better), with distributions capturing dataset-wise variability.
Annotations $(D)$ indicate the number of datasets evaluated per method; \BEACON and the task-specific \HazeMatching/\ResMatching baselines cover all $10$ datasets. \BEACON ($K=16$, bold black) achieves the lowest median rank across all four metric pairs and a lower mean rank than the guided flow-matching baselines in three of the four, with a marginally higher mean rank on MicroMS-SSIM--FID.
Complete dataset-wise results are provided in Supp.\ Section~\ref{supp:results}.}
  \label{fig:rank_main}
\end{figure*}
}

\newcommand{\figCalibMain}{
\begin{figure}[tb]
  \centering
  \includegraphics[width=\linewidth]{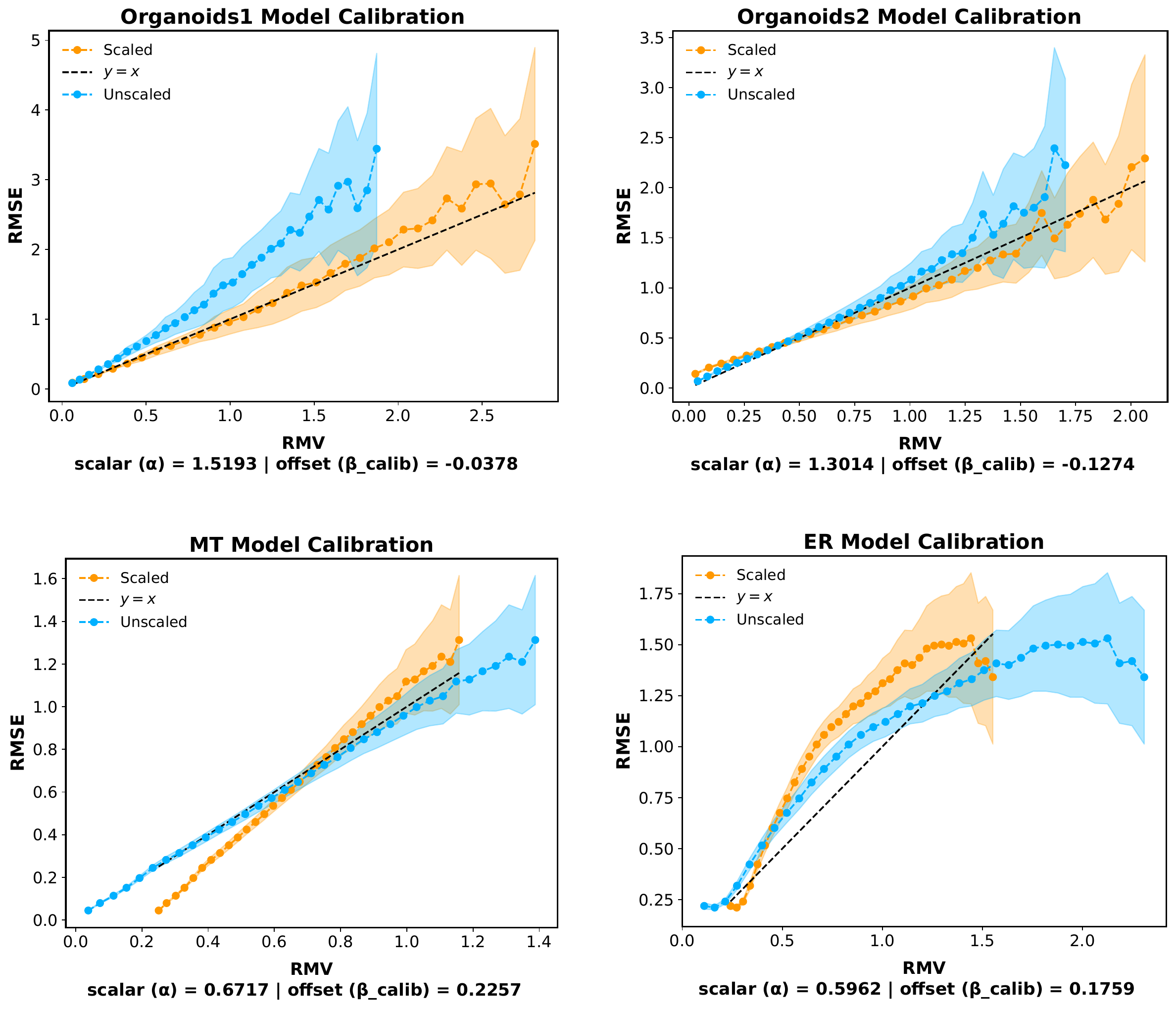}
  \caption{\textbf{Uncertainty calibration of \BEACON ($K=16$) across two restoration tasks.}
  The top row shows microscopy dehazing on \textit{Organoids1} (left) and \textit{Organoids2} (right); the bottom row shows super-resolution on \textit{MT} (left) and \textit{ER} (right).
  Following Section~\ref{sec:calibration}, pixels are grouped by predicted uncertainty, and each point compares the root mean variance (RMV, horizontal axis) of posterior predictions with the observed root mean squared error (RMSE, vertical axis) in a bin.
  Blue curves show unscaled uncertainty estimates, and orange curves show estimates after linear calibration (``Scaled''); the fitted scale and offset are printed below each panel.
  The black dashed identity line ($y=x$) indicates agreement between predicted uncertainty and reconstruction error: points above it indicate underestimated uncertainty, while points below it indicate overestimated uncertainty.
  Calibration brings the organoid curves closer to the identity line over much of the displayed range, while residual deviations remain for ER and MT, showing that a global linear correction does not ensure agreement in every uncertainty bin.
  }
  \label{fig:calib_main}
\end{figure}
}

\newcommand{\figQualitative}{
\begin{figure*}[h]
\centering
  \includegraphics[width=0.93\textwidth]{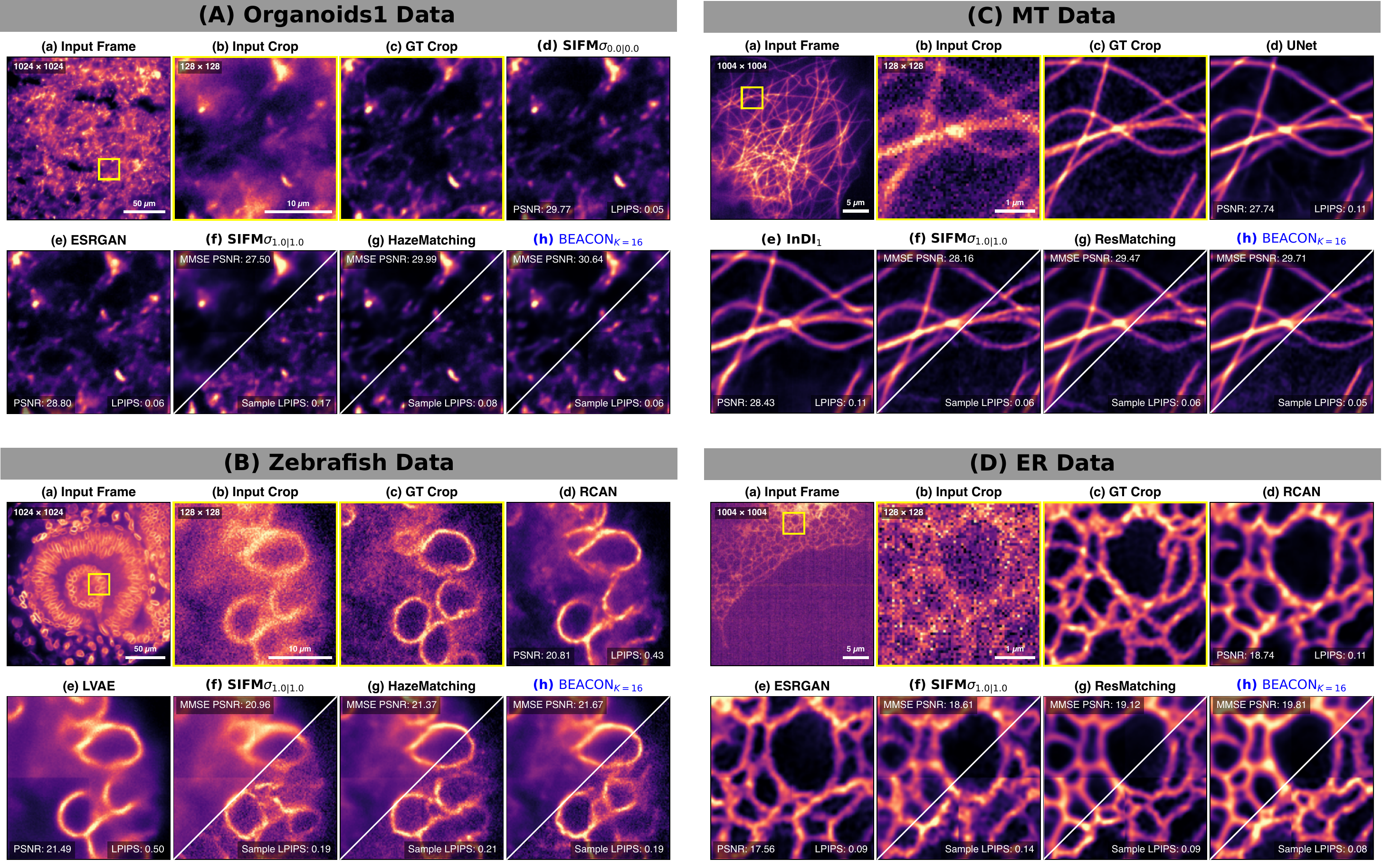}
    \caption{\textbf{Qualitative results.}
    We show representative restoration results for four datasets: \textbf{(A)} Organoids1, \textbf{(B)} Zebrafish, \textbf{(C)} Microtubules (MT, super-resolution), and \textbf{(D)} Endoplasmic Reticulum (ER).
    For each dataset, we show \textbf{(a)} the full degraded input ($1024\times1024$ or $1004\times1004$) with a selected crop ($128\times128$, yellow box), \textbf{(b)} the degraded crop, \textbf{(c)} the corresponding ground truth, \textbf{(d--g)} predictions from the four best competing methods by median perception--distortion rank (see Section~\ref{sec:evaluation_metrics}), and \textbf{(h)} \BEACON.
    For generative posterior models, diagonally split images show a representative posterior sample and the corresponding MMSE estimate, with sample LPIPS and MMSE PSNR reported, respectively.
    For point-predicting methods, PSNR and LPIPS are reported directly. 
    Additional results are provided in Supp.\ Section~\ref{supp:qual_results}.}
  \label{fig:qualitative}
\end{figure*}
}

\newcommand{\figQuantitativeOne}{
\begin{figure*}[h]
\centering
\includegraphics[width=\textwidth]{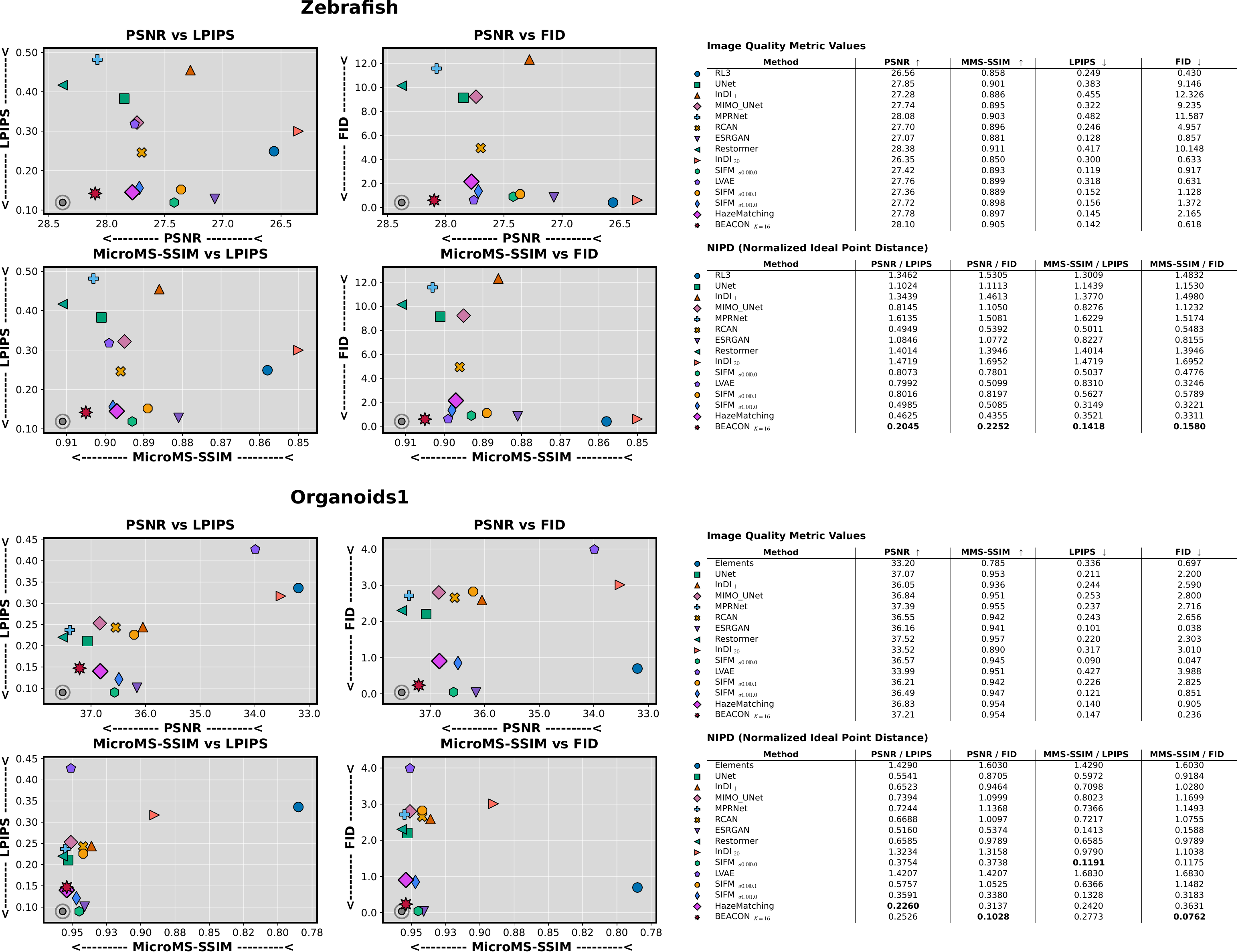}
\caption{\textbf{Perception--distortion trade-off on Zebrafish and Organoids1.}
We compare \BEACON ($K=16$) against competing restoration methods on
\textbf{(A)} Zebrafish and \textbf{(B)} Organoids1.
For each dataset, we show PSNR vs.\ LPIPS (top-left),
PSNR vs.\ FID (top-right), MicroMS-SSIM vs.\ LPIPS
(bottom-left), and MicroMS-SSIM vs.\ FID (bottom-right).
Higher PSNR and MicroMS-SSIM indicate better fidelity,
while lower LPIPS and FID indicate better perceptual quality.
The empirical ideal point, combining the best observed
distortion and perceptual scores, is indicated by
\scalebox{1.8}{\(\textcolor{gray}{\odot}\)},
while colored markers denote individual methods.
Axes are oriented such that the bottom-left corner represents
the most favorable trade-off.
Corresponding numerical results along with NIPD scores (best one in \textbf{bold}) are provided on the right.}
\label{fig:quantitative1}
\end{figure*}
}

\newcommand{\figQuantitativeTwo}{
\begin{figure*}[tbp]
\centering
\includegraphics[width=\textwidth]{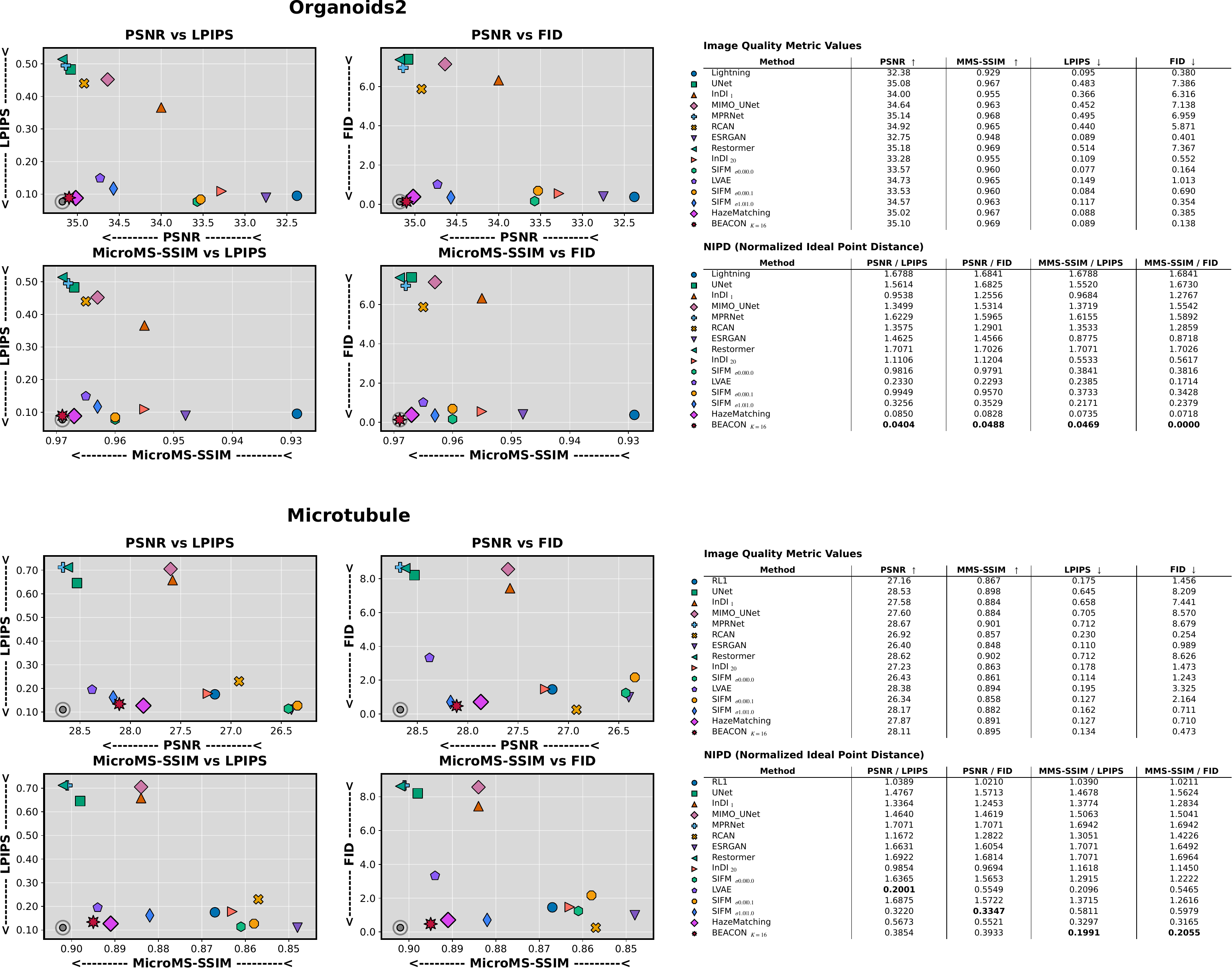}
\caption{\textbf{Perception--distortion trade-off on Organoids2 and Microtubule.}
We compare \BEACON ($K=16$) against competing restoration methods on
\textbf{(A)} Organoids2 and \textbf{(B)} Microtubule (dehazing; distinct from the super-resolution MT dataset).
For each dataset, four metric combinations are shown:
PSNR vs.\ LPIPS, PSNR vs.\ FID,
MicroMS-SSIM vs.\ LPIPS, and MicroMS-SSIM vs.\ FID.
Higher PSNR and MicroMS-SSIM indicate better fidelity,
while lower LPIPS and FID indicate better perceptual quality.
The empirical ideal point, combining the best observed
distortion and perceptual scores, is indicated by
\scalebox{1.8}{\(\textcolor{gray}{\odot}\)},
while colored markers denote individual methods.
Axes are oriented such that the bottom-left corner represents
the most favorable trade-off.
Corresponding numerical results along with NIPD scores (best one in \textbf{bold}) are provided on the right..}
\label{fig:quantitative2}
\end{figure*}
}

\newcommand{\figQuantitativeThree}{
\begin{figure*}[tbp]
\centering
\includegraphics[width=\textwidth]{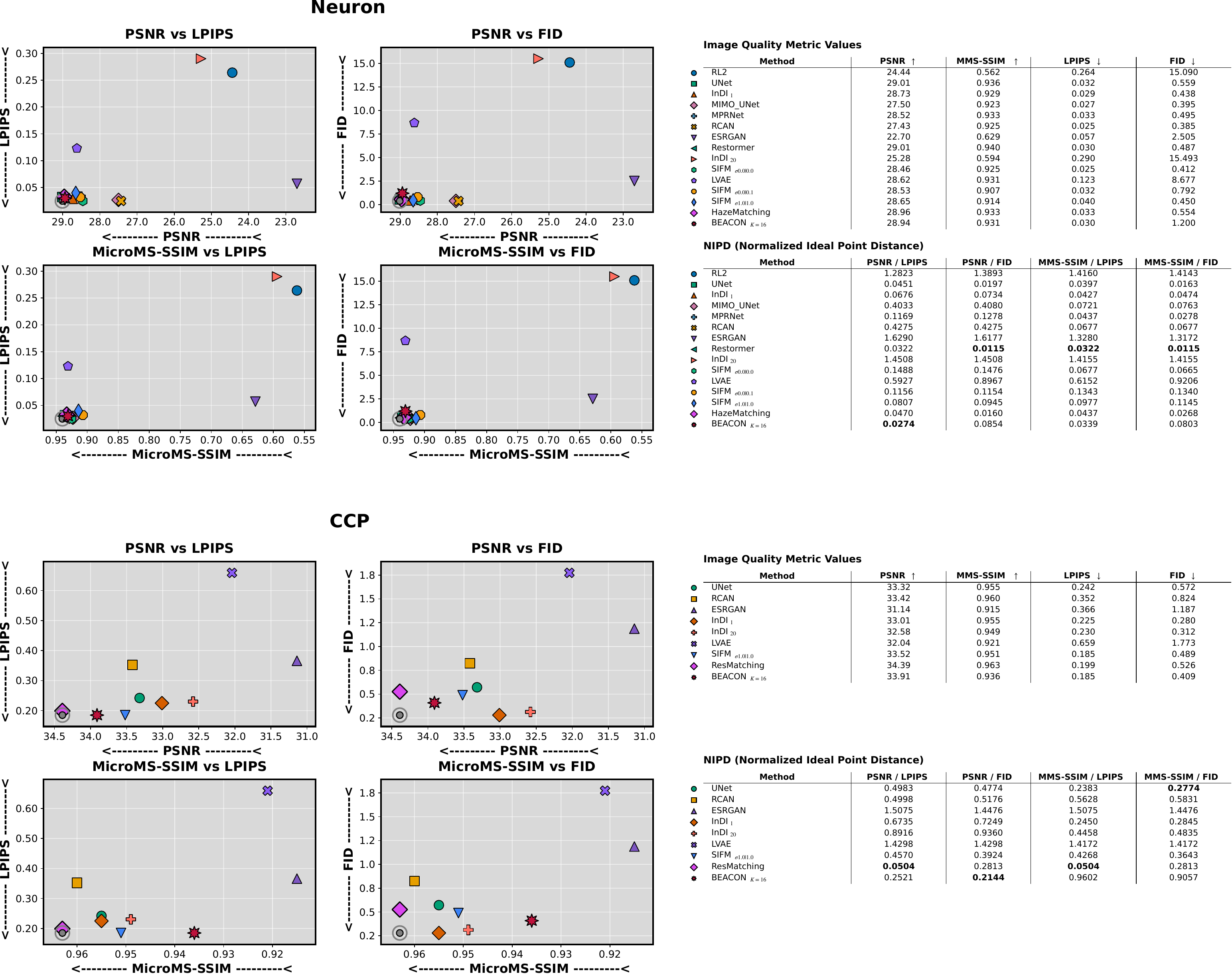}
\caption{\textbf{Perception--distortion trade-off on Neurons and CCP.}
We compare \BEACON ($K=16$) against competing restoration methods on
\textbf{(A)} Neurons and \textbf{(B)} Clathrin-Coated Pits (CCP).
For each dataset, we show PSNR vs.\ LPIPS (top-left),
PSNR vs.\ FID (top-right), MicroMS-SSIM vs.\ LPIPS
(bottom-left), and MicroMS-SSIM vs.\ FID (bottom-right).
Higher PSNR and MicroMS-SSIM indicate better fidelity,
while lower LPIPS and FID indicate better perceptual quality.
The empirical ideal point, combining the best observed
distortion and perceptual scores, is indicated by
\scalebox{1.8}{\(\textcolor{gray}{\odot}\)},
while colored markers denote individual methods.
Axes are oriented such that the bottom-left corner represents
the most favorable trade-off.
Corresponding numerical results along with NIPD scores (best one in \textbf{bold}) are provided on the right.}
\label{fig:quantitative3}
\end{figure*}
}

\newcommand{\figQuantitativeFour}{
\begin{figure*}[tbp]
\centering
\includegraphics[width=\textwidth]{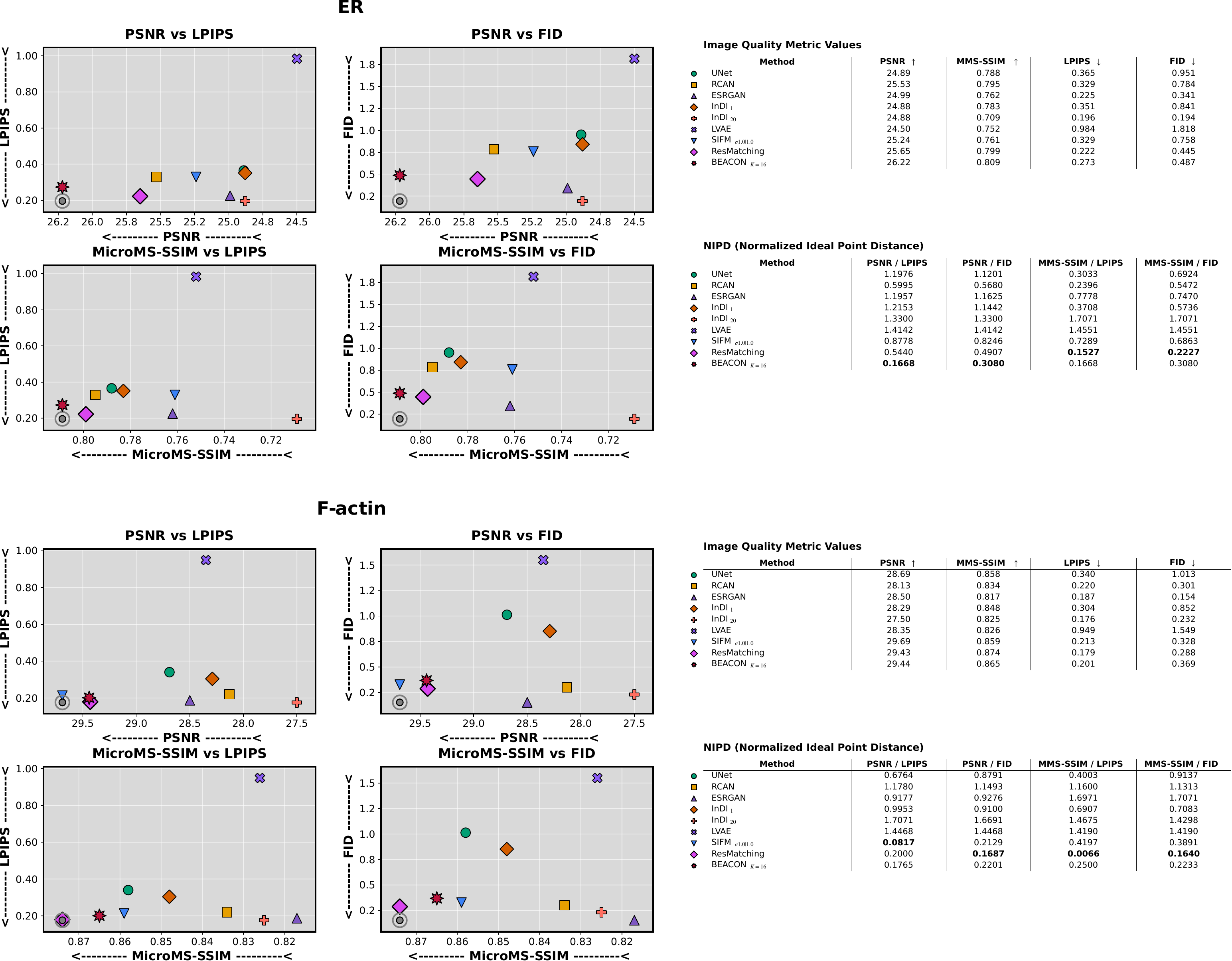}
\caption{\textbf{Perception--distortion trade-off on ER and F-actin.}
We compare \BEACON ($K=16$) against competing restoration methods on
\textbf{(A)} Endoplasmic Reticulum (ER) and \textbf{(B)} F-actin.
For each dataset, we show PSNR vs.\ LPIPS (top-left),
PSNR vs.\ FID (top-right), MicroMS-SSIM vs.\ LPIPS
(bottom-left), and MicroMS-SSIM vs.\ FID (bottom-right).
Higher PSNR and MicroMS-SSIM indicate better fidelity,
while lower LPIPS and FID indicate better perceptual quality.
The empirical ideal point, combining the best observed
distortion and perceptual scores, is indicated by
\scalebox{1.8}{\(\textcolor{gray}{\odot}\)},
while colored markers denote individual methods.
Axes are oriented such that the bottom-left corner represents
the most favorable trade-off.
Corresponding numerical results along with NIPD scores (best one in \textbf{bold}) are provided on the right.}
\label{fig:quantitative4}
\end{figure*}
}

\newcommand{\figQuantitativeFive}{
\begin{figure*}[tbp]
\centering
\includegraphics[width=\textwidth]{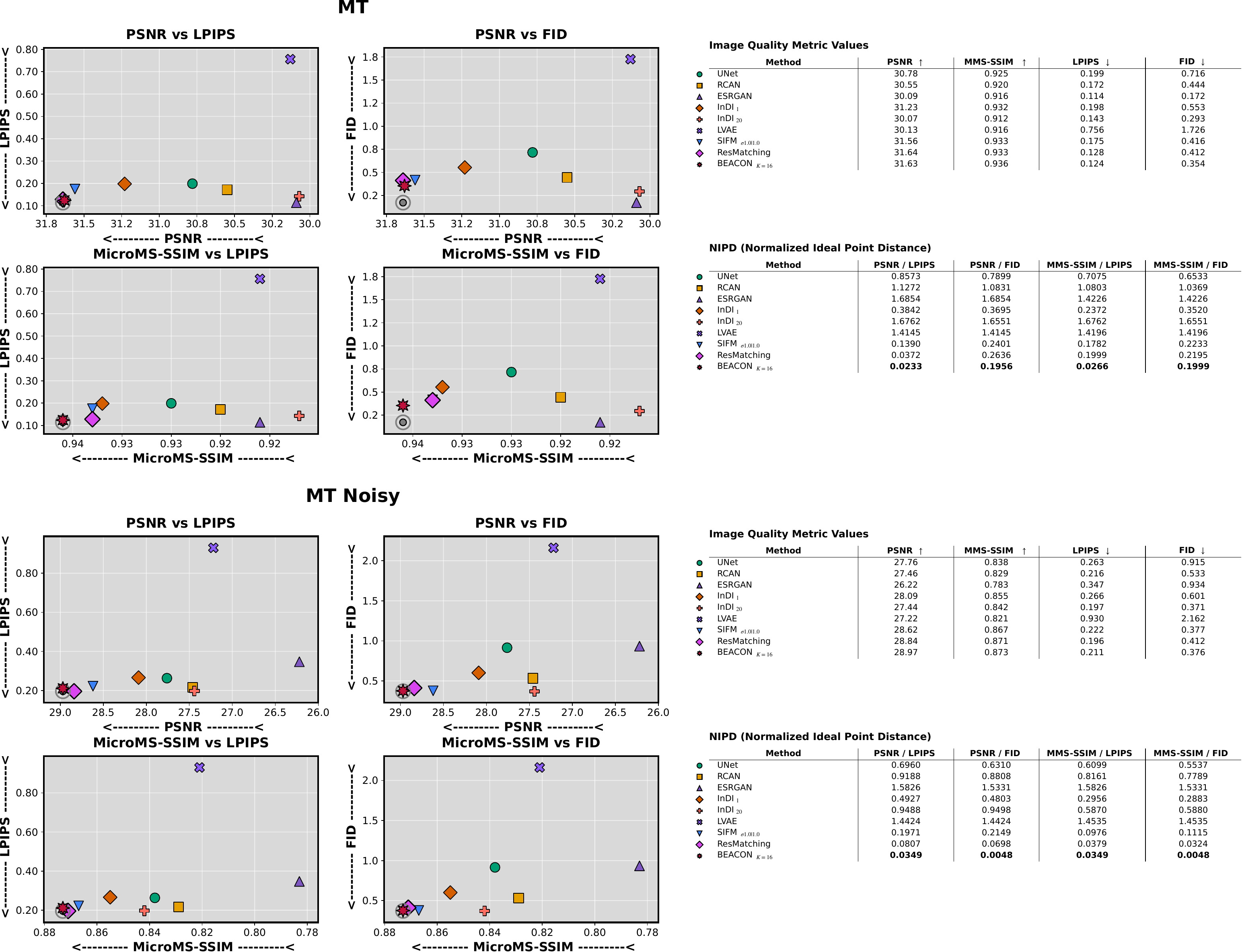}
\caption{\textbf{Perception--distortion trade-off on MT and MT-Noisy.}
We compare \BEACON ($K=16$) against competing restoration methods on
\textbf{(A)} Microtubules (MT) and
\textbf{(B)} Noisy Microtubules (MT-Noisy).
For each dataset, we show PSNR vs.\ LPIPS (top-left),
PSNR vs.\ FID (top-right), MicroMS-SSIM vs.\ LPIPS
(bottom-left), and MicroMS-SSIM vs.\ FID (bottom-right).
Higher PSNR and MicroMS-SSIM indicate better fidelity,
while lower LPIPS and FID indicate better perceptual quality.
The empirical ideal point, combining the best observed
distortion and perceptual scores, is indicated by
\scalebox{1.8}{\(\textcolor{gray}{\odot}\)},
while colored markers denote individual methods.
Axes are oriented such that the bottom-left corner represents
the most favorable trade-off.
Corresponding numerical results along with NIPD scores (best one in \textbf{bold}) are provided on the right.}
\label{fig:quantitative5}
\end{figure*}
}

\newcommand{\figAblationK}{
\begin{figure*}[h]
\centering
\includegraphics[width=\textwidth]{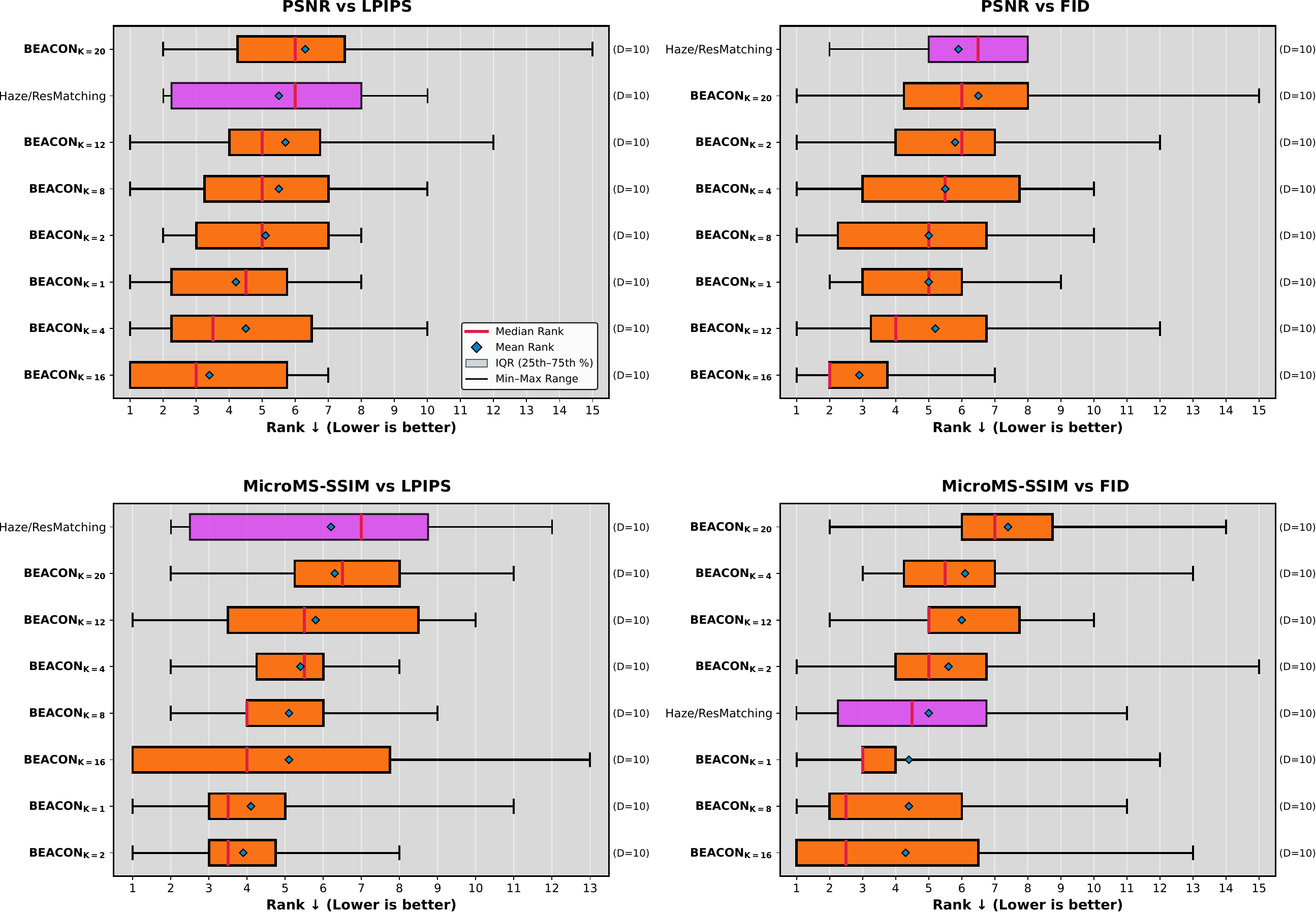}
\caption{\textbf{Rank distributions of \BEACON variants across the ten test datasets.}
We evaluate the effect of varying the number of noise candidates
($K \in \{1,2,4,8,12,16,20\}$) on \BEACON's
perception--distortion performance across four metric pairs:
PSNR--LPIPS (top-left), PSNR--FID (top-right),
MicroMS-SSIM--LPIPS (bottom-left), and MicroMS-SSIM--FID
(bottom-right), using LPIPS (Ind) and FID (Ind).
For each dataset, the empirical ideal point used to compute
the NIPD score is determined from all available methods,
including competing baselines and all \BEACON variants
evaluated here. Consequently, the displayed \BEACON variants
retain their ranks relative to the full set of competing methods,
rather than being ranked exclusively against one another
(see Section~\ref{sec:evaluation_metrics}).
Each box summarizes the ranks across the ten test datasets.
The annotation $D$ indicates the number of contributing datasets.
Variants are ordered by median rank, with mean rank used
to break ties. Lower ranks indicate a more favorable
perception--distortion trade-off, while narrower boxes
reflect greater consistency across datasets. For reference, we have shown how \HazeMatching/\ResMatching compares with different variants of \BEACON.}
\label{fig:ablation_k}
\end{figure*}
}

\newcommand{\figAblationBestK}{
\begin{figure*}[h]
\centering
\includegraphics[width=\textwidth]{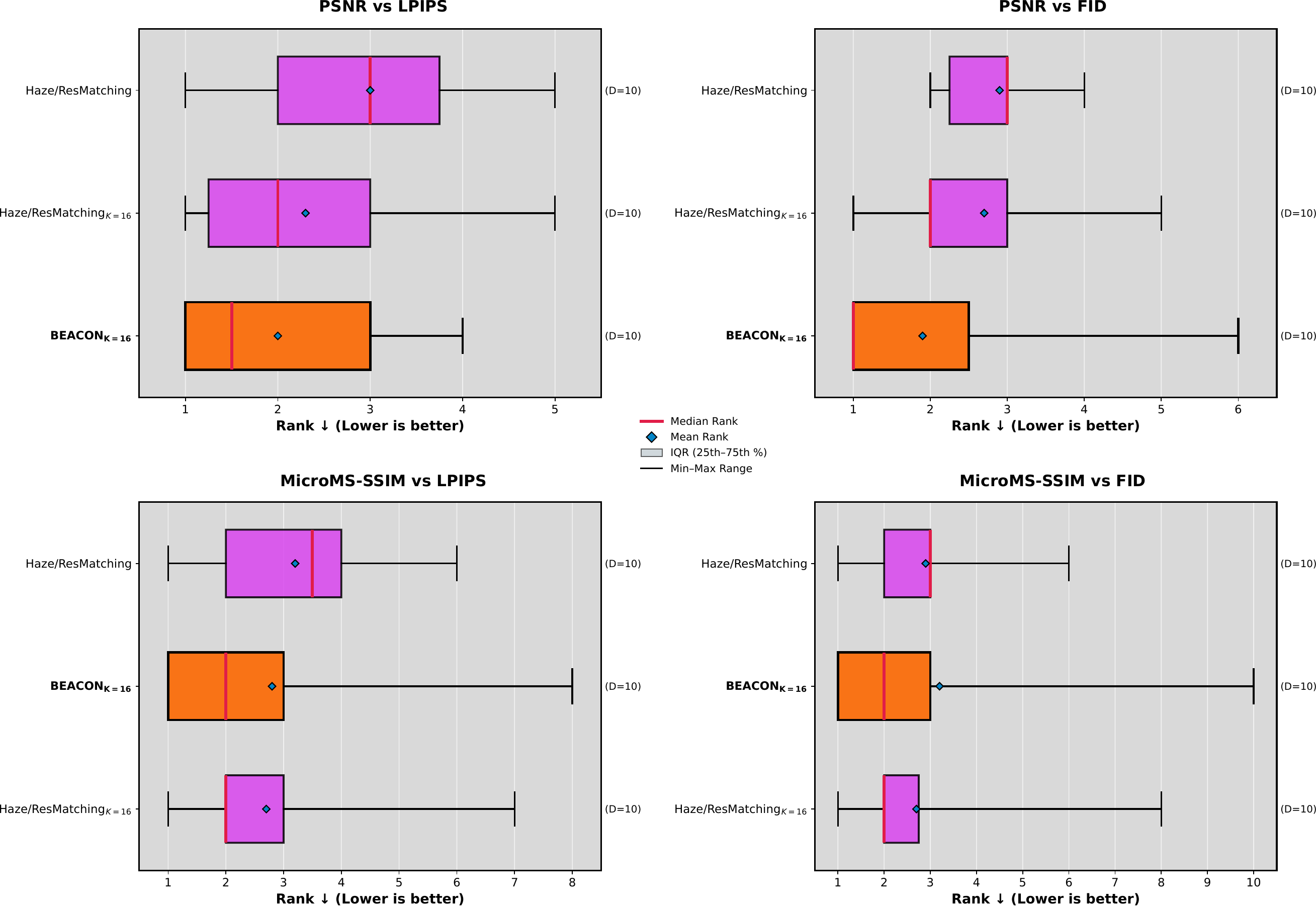}
\caption{\textbf{Effect of Best-of-$K$ training on \HazeMatching/\ResMatching baselines.}
We investigate whether the improvements achieved by \BEACON
can be attributed solely to Best-of-$K$ candidate selection
by incorporating the same training strategy ($K=16$)
into the original HazeMatching and ResMatching baselines,
while retaining their standard Gaussian noise initialization.
We compare the original \HazeMatching/\ResMatching with Best-of-$K$ ($K=16$), and \BEACON ($K=16$)
across four perception--distortion metric pairs:
PSNR--LPIPS (top-left), PSNR--FID (top-right),
MicroMS-SSIM--LPIPS (bottom-left), and MicroMS-SSIM--FID
(bottom-right), using LPIPS (Ind) and FID (Ind).
For each dataset, the empirical ideal point used to compute
the NIPD score is determined from all available methods,
including competing baselines and \BEACON ($K=16$) variant.
Consequently, the displayed methods retain their ranks
relative to the full set of competing methods,
rather than being ranked exclusively against one another
(see Section~\ref{sec:evaluation_metrics}).
Each box summarizes the ranks across the ten test datasets.
The annotation $D$ indicates the number of contributing datasets.
Methods are ordered by median rank, with mean rank used
to break ties. Lower ranks indicate a more favorable
perception--distortion trade-off, while narrower boxes
reflect greater consistency across datasets.
\BEACON achieves lower median ranks than the Best-of-$K$
baselines on both PSNR-based metric pairs and matches their
median ranks on both MicroMS-SSIM-based pairs, although its
mean ranks are slightly higher on the latter.
These results suggest that learning an observation-conditioned
noise distribution provides additional benefits beyond
Best-of-$K$ selection for PSNR-based perception--distortion
trade-offs, while improvements are not uniform across metrics.}
\label{fig:ablation_best_k}
\end{figure*}
}

\newcommand{\figNoiseanalysis}{
\begin{figure*}[t]
    \centering
    \includegraphics[width=0.98\textwidth]{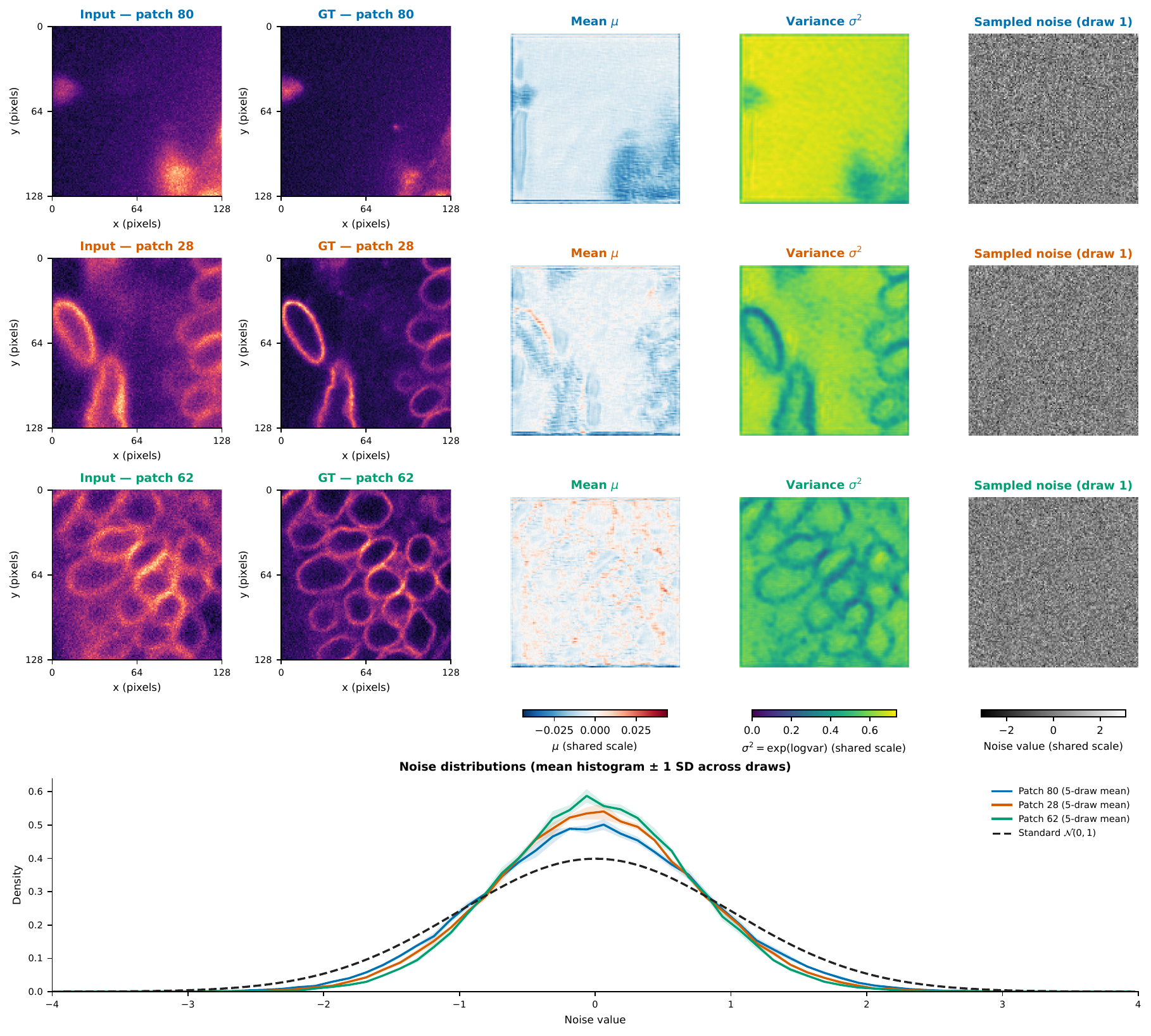}
    \caption{
    \textbf{Spatially adaptive noise distributions learned by \BEACON.}
    Visualization of the learned conditional noise model on
    three representative patches from the Zebrafish dataset
    with $K=16$.
    Each row shows the degraded observation
    $\mathbf{x}_{M_0}$, the corresponding ground truth
    $\mathbf{x}_{M_1}$, the predicted mean $\boldsymbol{\mu}$,
    the predicted variance $\boldsymbol{\sigma}^2$,
    and one sampled noise realization.
    Although individual noise samples appear visually
    similar to random Gaussian noise, the predicted
    variance maps exhibit spatial structure correlated
    with the underlying morphology, assigning lower
    variance to prominent cellular structures, while the
    predicted mean stays close to zero (note its scale).
    The learned proposal thus spatially modulates the
    noise scale.
    The bottom panel compares the empirical distributions
    of sampled noise (averaged over five draws, with
    $\pm 1$ standard deviation) against the standard
    Gaussian $\mathcal{N}(0,I)$.
    The learned noise distributions are more concentrated
    around zero, illustrating how \BEACON departs from
    an isotropic Gaussian prior to construct
    observation-dependent initializations for conditional
    flow matching.
    }
    \label{fig:conditional_noise_visualization}
\end{figure*}
}

\newcommand{\figValKSelection}{
\begin{figure*}[tb]
    \centering
    \includegraphics[width=0.8\textwidth]{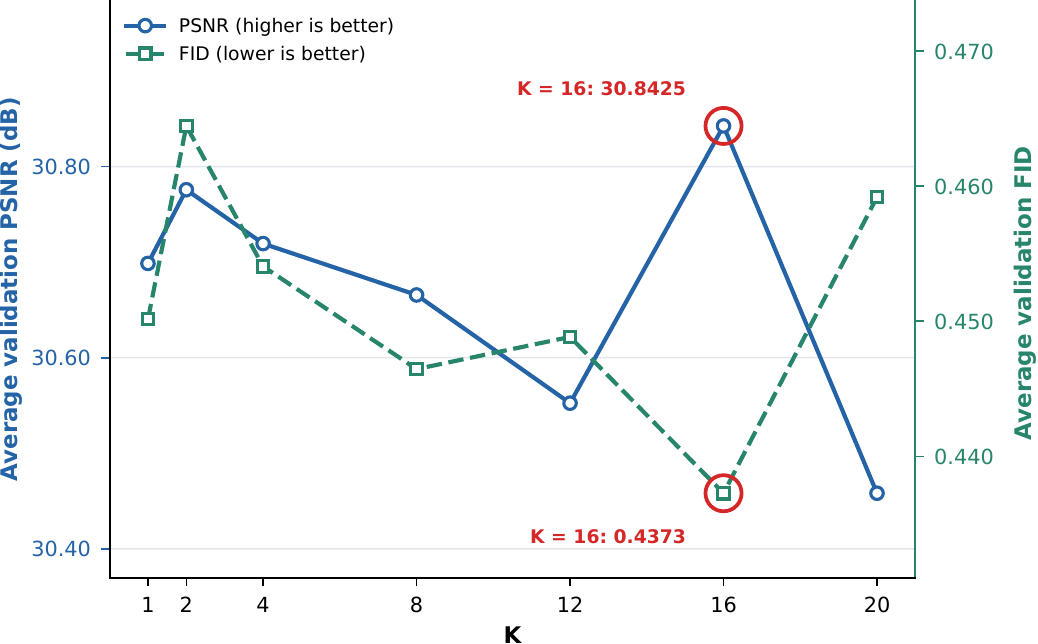}
    \caption{
    \textbf{Effect of the number of candidates ($K$).}
    Average validation PSNR (higher is better) and FID
    (lower is better) across all 10 datasets for different
    values of $K \in \{1,2,4,8,12,16,20\}$.
    Both metrics achieve their best average values at
    $K=16$, with a PSNR of $30.84$\,dB and an FID
    of $0.4373$.
    Differences between values of $K$ are small, and
    neither metric varies monotonically with $K$.
    We therefore select $K=16$ for all experiments
    reported in the main paper.
    }
    \label{fig:supp_k_ablation}
\end{figure*}
}

\newcommand{\figCalibEfficiency}{
\begin{figure*}[b]
    \centering
    \includegraphics[width=0.91\textwidth]{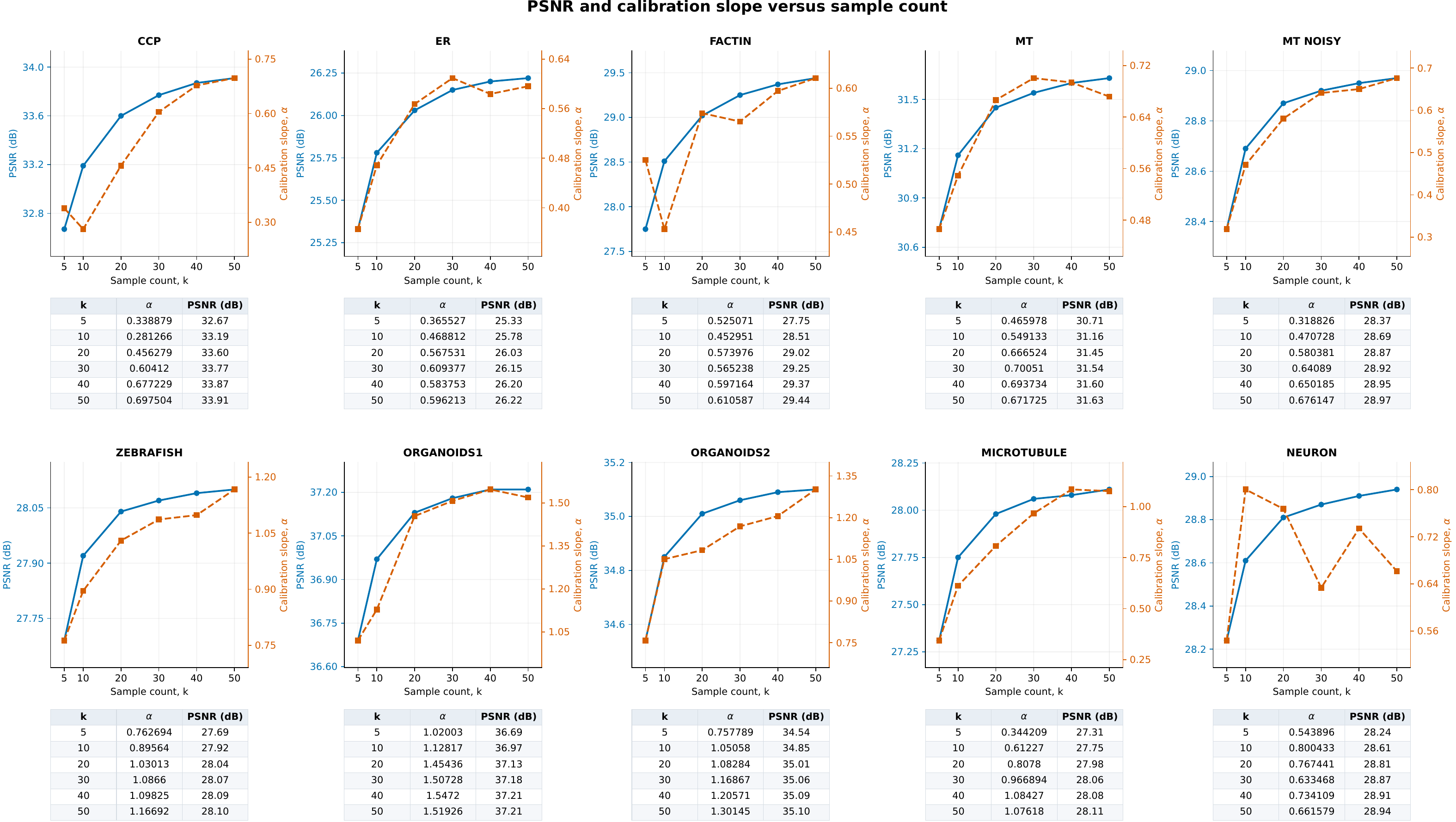}
    \caption{
    \textbf{Effect of posterior sample count on reconstruction
    fidelity and uncertainty calibration.}
    We evaluate the influence of the number of posterior
    samples ($k \in \{5,10,20,30,40,50\}$) across all
    ten microscopy datasets. Each panel reports the
    peak signal-to-noise ratio (PSNR, blue solid line)
    and the uncertainty calibration slope
    ($\alpha$, orange dashed line) as functions of
    $k$. PSNR is computed from the mean of the
    posterior samples, while $\alpha$ is obtained
    from the linear relationship between predicted
    uncertainty and reconstruction error.
    Increasing the number of samples consistently
    improves PSNR, with diminishing returns
    beyond approximately 20--30 samples.
    The calibration slope exhibits greater
    dataset-dependent variability, generally
    becoming more stable at larger sample counts.
    These results illustrate the trade-off between
    posterior sampling cost, reconstruction fidelity,
    and uncertainty estimation.
    }
    \label{fig:calib_efficiency}
\end{figure*}
}

\newcommand{\figCalibrationSupp}{
\begin{figure*}[h]
\centering
  \includegraphics[width=\textwidth]{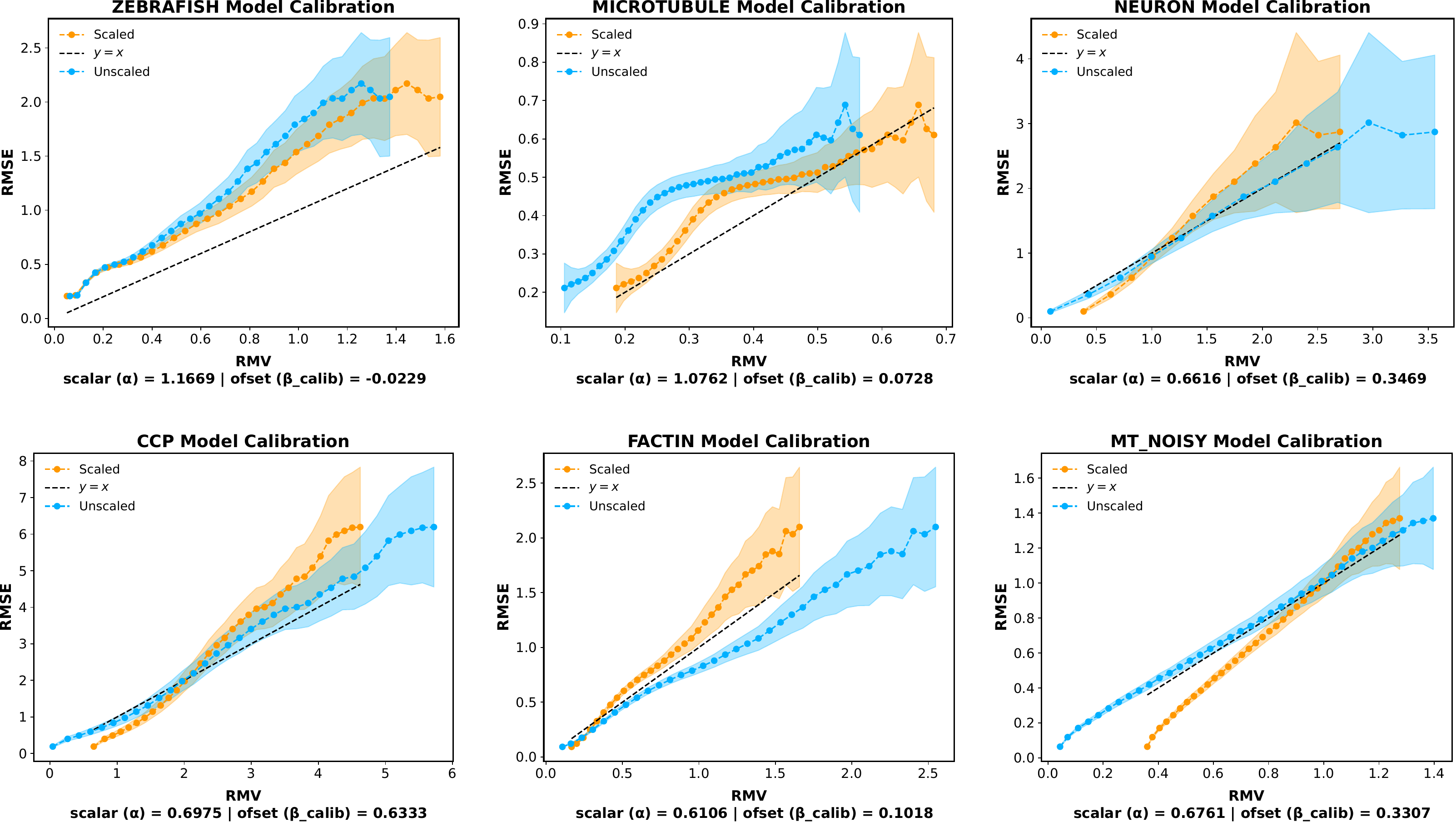}
  \caption{\textbf{Uncertainty calibration across six microscopy datasets.}
    Comparison of the root mean variance (RMV),
    estimated from the pixel-wise variability of posterior samples,
    against the root mean squared error (RMSE) with respect to
    ground-truth images.
    Results are shown for Zebrafish, Microtubule, Neuron,
    CCP, F-actin, and MT-Noisy datasets.
    Blue curves show uncalibrated uncertainty estimates,
    while orange curves show estimates after applying a
    dataset-specific linear calibration,
    $\mathrm{RMV}_{\mathrm{cal}} =
    \alpha\,\mathrm{RMV}+\beta_{\mathrm{calib}}$.
    The fitted scaling factor $\alpha$ and offset
    $\beta_{\mathrm{calib}}$ are reported below each panel.
    The dashed black line ($y=x$) indicates perfect calibration,
    where estimated uncertainty agrees with the observed
    reconstruction error.
    Shaded regions indicate variability around the corresponding
    curves. Closer agreement with the diagonal indicates
    better uncertainty calibration.
  }
  \label{fig:calibration_supp}
\end{figure*}
}

\newcommand{\figTrainTimeK}{
\begin{figure}[h]
    \centering
  \includegraphics[width=0.8\textwidth]
  {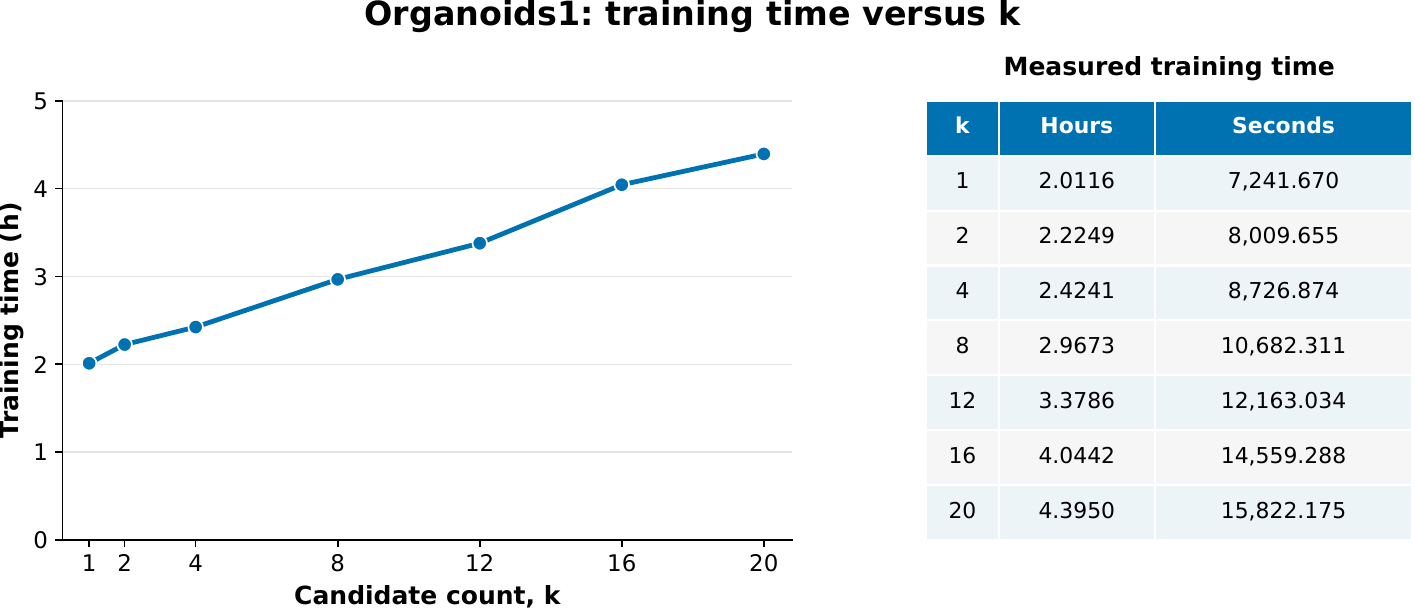}
    \caption{\textbf{Training-time overhead as a function of candidate count $K$, on the Organoids1 Dataset}
    Wall-clock training time (hours) for \BEACON with
    varying numbers of candidates in the best-of-$K$
    selection procedure. Training time increases
    gradually with $K$, from approximately 2 hours
    at $K=1$ to 4 hours at $K=20$,
    illustrating the computational overhead associated
    with exploring additional candidate couplings
    during training.}
    \label{fig:k_training_time}
\end{figure}
}

\newcommand{\figStepsPSNRLPIPS}{
\begin{figure*}[h]
    \centering
    \includegraphics[width=1.0\textwidth]{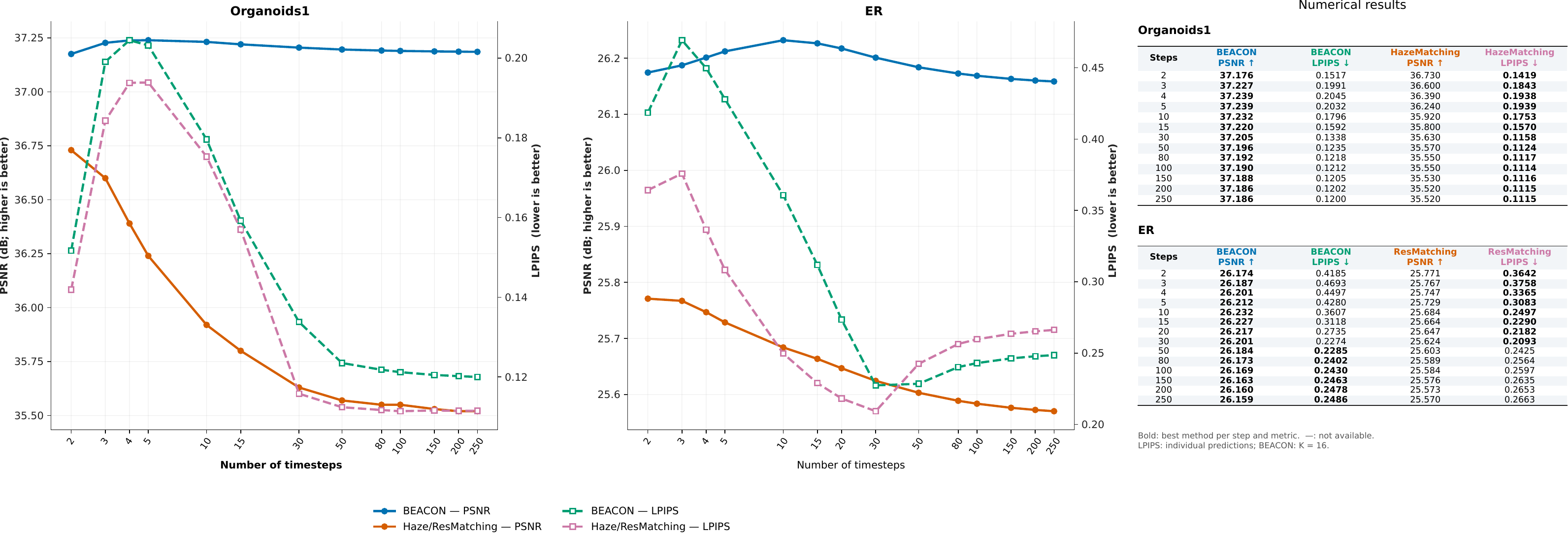}
    \caption{
    \textbf{Effect of inference timesteps on reconstruction
    fidelity and perceptual quality.}
    We evaluate the influence of the number of inference
    timesteps on \BEACON ($K=16$) and the corresponding
    \HazeMatching~\citep{ray2026hazematching} and \ResMatching~\citep{resmatching2026} baselines using
    Organoids1 (dehazing) and ER (computational
    super-resolution).
    Each panel reports PSNR (solid lines) and LPIPS (dashed lines) as functions 
    of the number of inference timesteps.
    \BEACON consistently maintains higher PSNR across
    both datasets, exhibiting substantially less
    degradation as the number of timesteps increases.
    While LPIPS varies with the integration step count,
    \BEACON achieves competitive perceptual quality,
    particularly at larger numbers of timesteps.
    These results demonstrate that \BEACON is less
    sensitive to the choice of inference timesteps,
    maintaining a favorable perception--distortion
    trade-off over a wide range of integration steps.
    }
    \label{fig:steps_psnr_lpips}
\end{figure*}
}

\newcommand{\figSamplesPSNR}{
\begin{figure*}[h]
    \centering
    \includegraphics[width=1.0\textwidth]{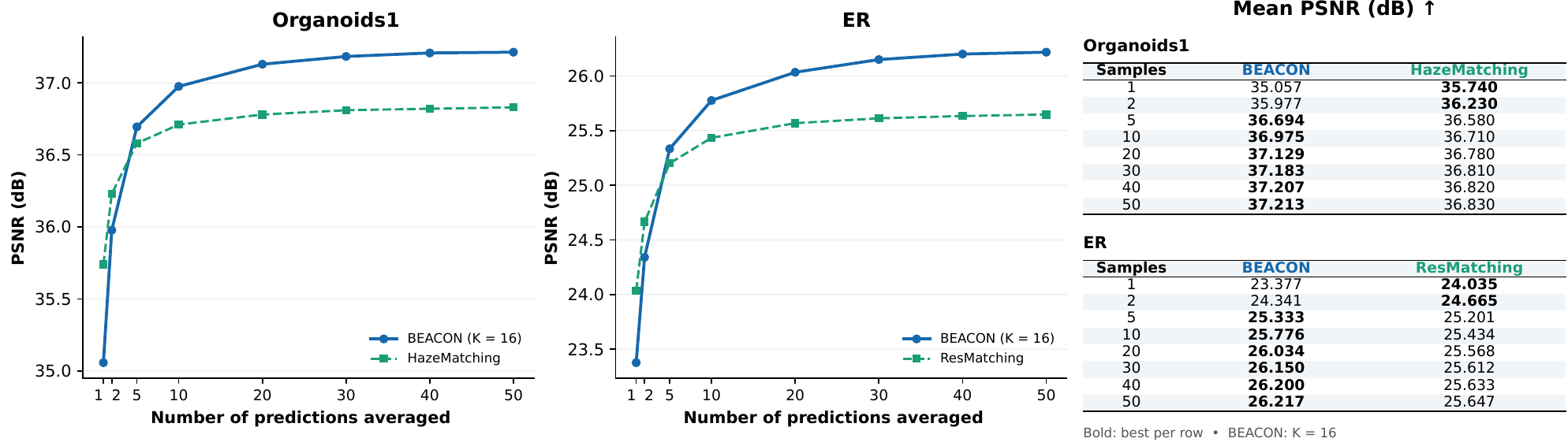}
    \caption{
    \textbf{Effect of posterior sample averaging on
    reconstruction fidelity.}
    We evaluate the influence of the number of posterior
    predictions averaged ($k \in \{1,2,5,10,20,30,40,50\}$)
    on \BEACON ($K=16$) and the corresponding \HazeMatching~\citep{ray2026hazematching} and \ResMatching~\citep{resmatching2026} baselines using Organoids1 (dehazing)
    and ER (computational super-resolution).
    Each panel reports the PSNR computed from the pixel-wise mean of $k$ posterior predictions.
    Increasing the number of averaged predictions
    improves PSNR for all methods, with diminishing
    returns beyond approximately 20--30 samples.
    Although \BEACON surpasses the corresponding baselines with as few as five averaged predictions and maintains
    a consistent advantage as $k$ increases.
    These results demonstrate that \BEACON benefits
    more strongly from posterior averaging, achieving
    higher reconstruction fidelity with relatively
    few posterior samples.
    }
    \label{fig:samples_psnr}
\end{figure*}
}

\newcommand{\figQualitativeSuppOne}{
\begin{figure*}[h]
\centering
  \includegraphics[width=0.8\textwidth]{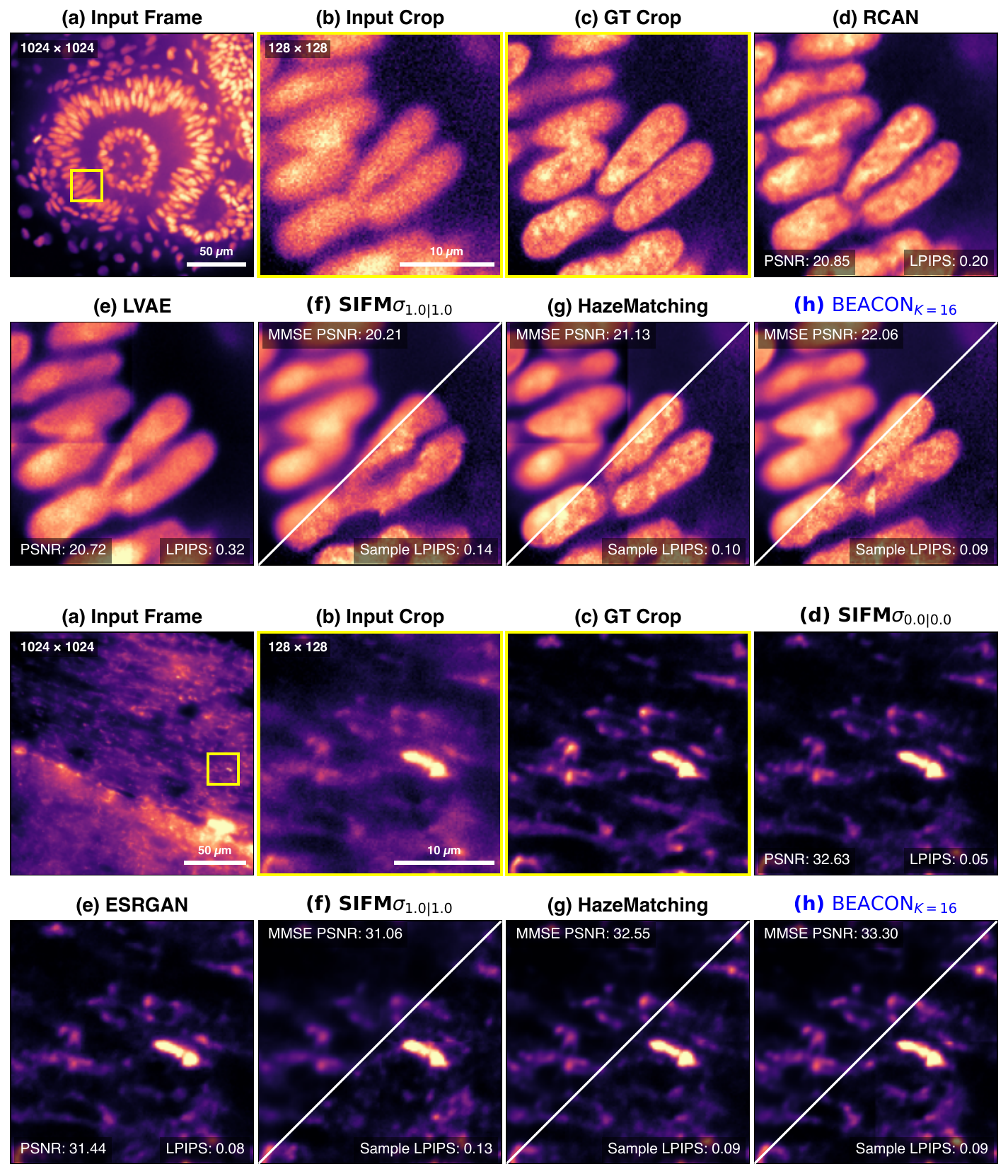}
    \caption{\textbf{Additional qualitative results.}
    We show representative restoration results for two datasets:
    Zebrafish (upper panels) and
    Organoids1 (lower panels).
    For each dataset, we show
    \textbf{(a)} the full degraded input with a selected crop (yellow box),
    \textbf{(b)} the degraded crop,
    \textbf{(c)} the corresponding ground truth,
    \textbf{(d--g)} predictions from four competing methods, and
    \textbf{(h)} \BEACON{} ($K=16$).
    For generative posterior models, diagonally split images show the MMSE estimate (upper-left) and a representative posterior sample (lower-right), with MMSE PSNR and sample LPIPS reported, respectively.
    For point-predicting methods, PSNR and LPIPS are reported directly.
}
  \label{fig:qualitative_supp1}
\end{figure*}
}

\newcommand{\figQualitativeSuppTwo}{
\begin{figure*}[t]
\centering
  \includegraphics[width=0.8\textwidth]{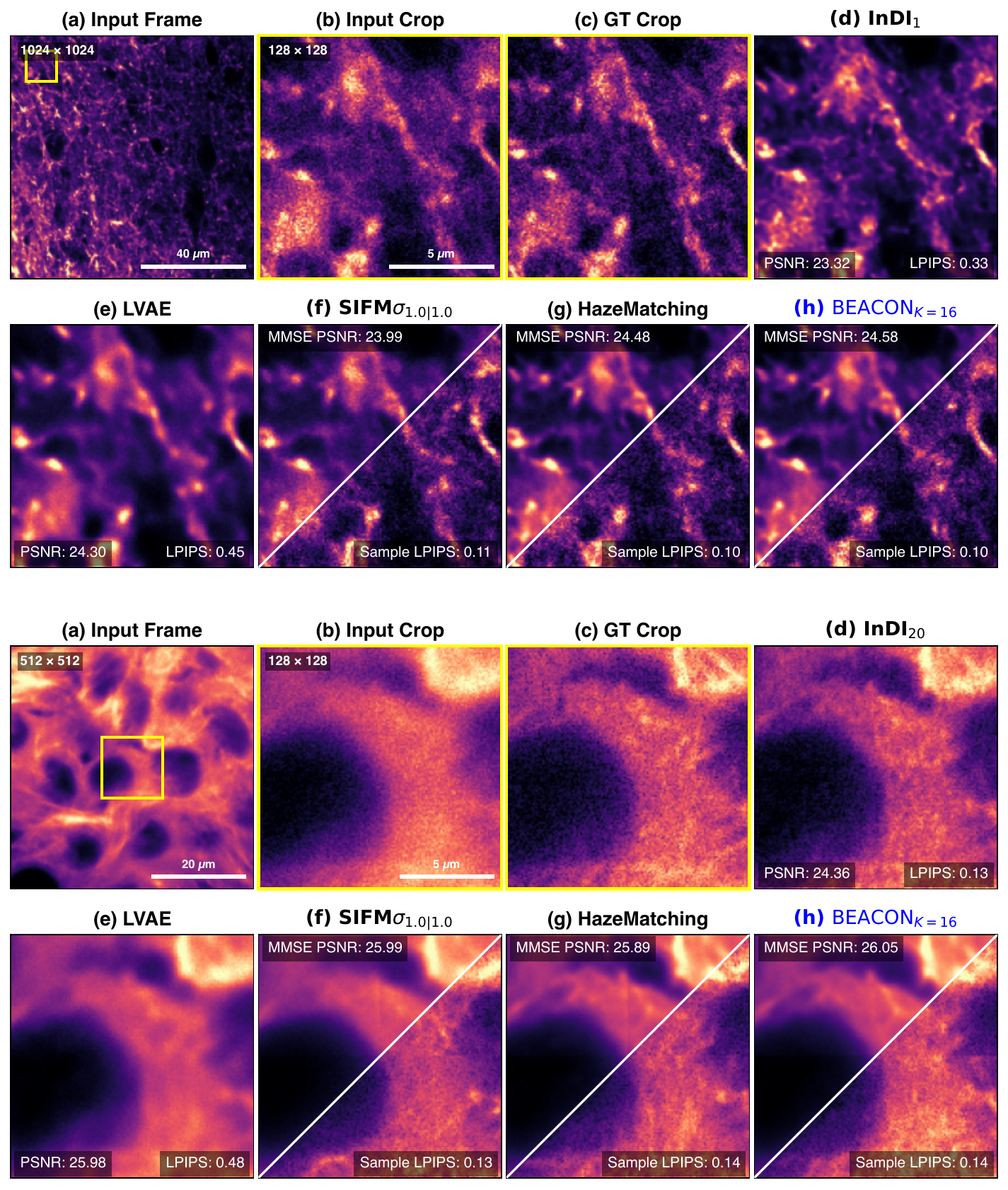}
    \caption{\textbf{Additional qualitative results.}
    We show representative restoration results for two datasets:
    Organoids2 (upper panels) and
    Microtubule (lower panels).
    For each dataset, we show
    \textbf{(a)} the full degraded input with a selected crop (yellow box),
    \textbf{(b)} the degraded crop,
    \textbf{(c)} the corresponding ground truth,
    \textbf{(d--g)} predictions from four competing methods, and
    \textbf{(h)} \BEACON{} ($K=16$).
    For generative posterior models, diagonally split images show the MMSE estimate (upper-left) and a representative posterior sample (lower-right), with MMSE PSNR and sample LPIPS reported, respectively.
    For point-predicting methods, PSNR and LPIPS are reported directly.
}
  \label{fig:qualitative_supp2}
\end{figure*}
}

\newcommand{\figQualitativeSuppThree}{
\begin{figure*}[t]
\centering
  \includegraphics[width=0.8\textwidth]{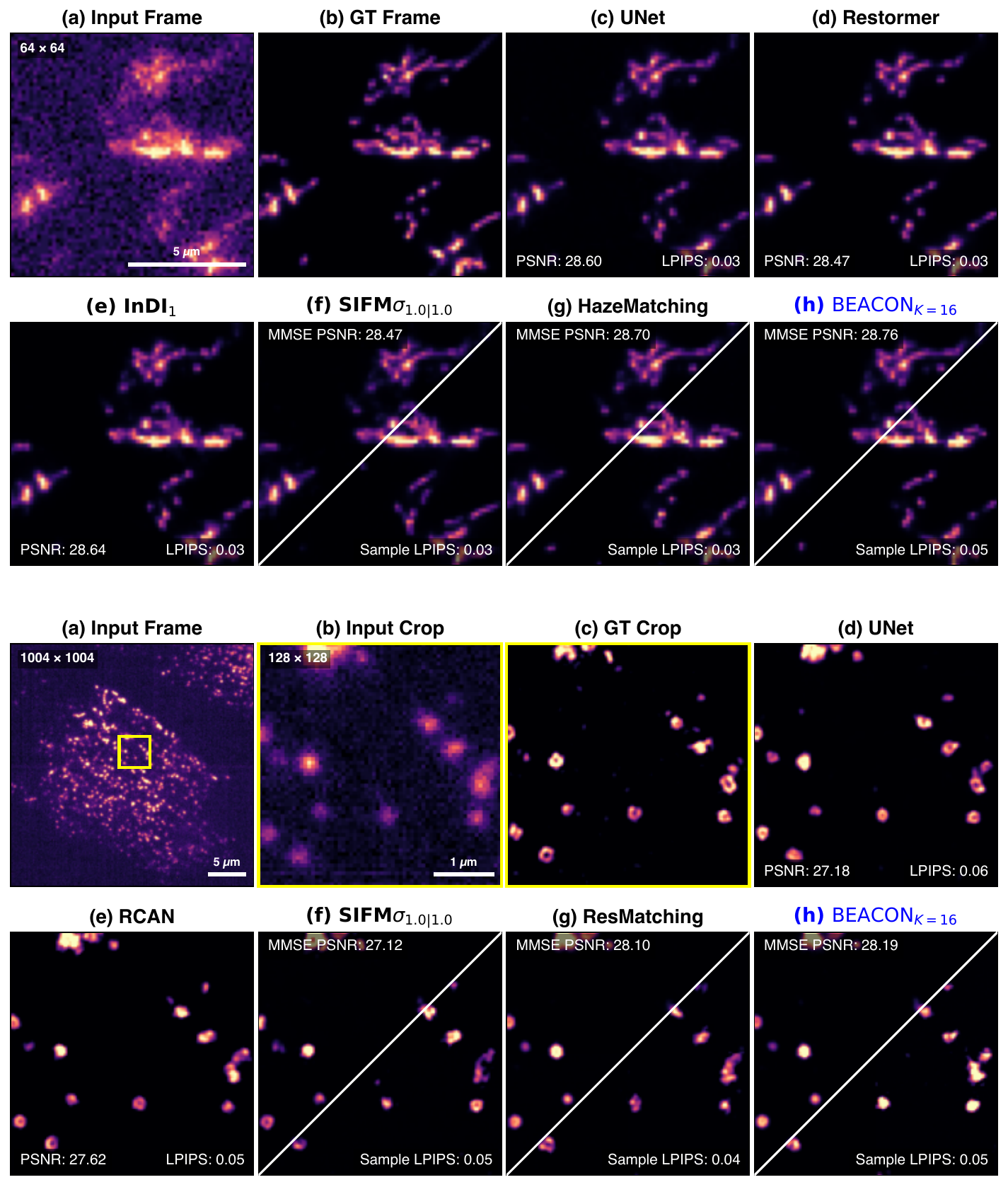}
    \caption{\textbf{Additional qualitative results.}
    We show representative restoration results for two datasets:
    Neuron (upper panels) and
    CCP (lower panels).
    For each dataset, we show
    \textbf{(a)} the full degraded input with a selected crop (yellow box),
    \textbf{(b)} the degraded crop,
    \textbf{(c)} the corresponding ground truth,
    \textbf{(d--g)} predictions from four competing methods, and
    \textbf{(h)} \BEACON{} ($K=16$).
    For generative posterior models, diagonally split images show the MMSE estimate (upper-left) and a representative posterior sample (lower-right), with MMSE PSNR and sample LPIPS reported, respectively.
    For point-predicting methods, PSNR and LPIPS are reported directly.
}
  \label{fig:qualitative_supp3}
\end{figure*}
}

\newcommand{\figQualitativeSuppFour}{
\begin{figure*}[t]
\centering
  \includegraphics[width=0.8\textwidth]{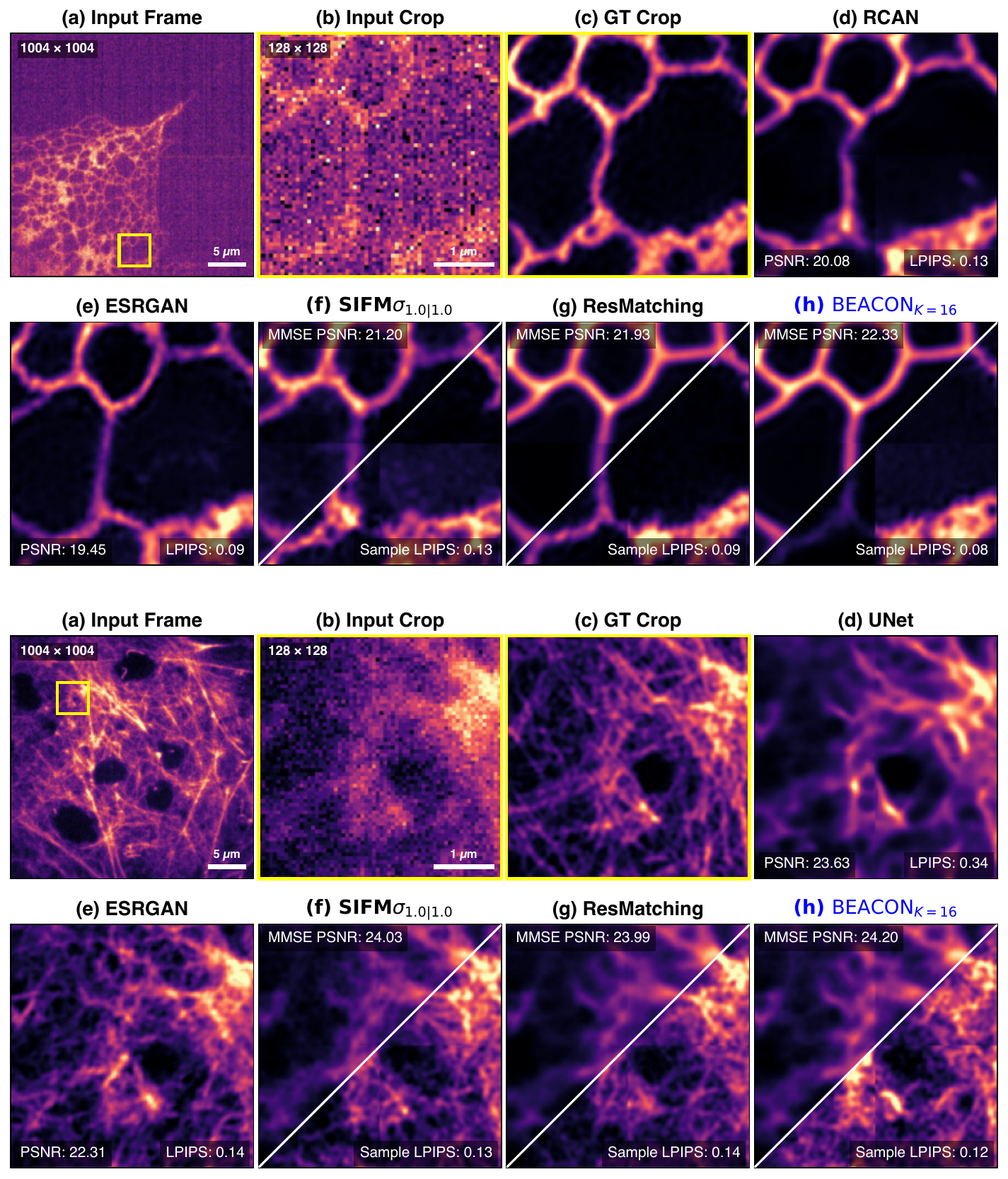}
    \caption{\textbf{Additional qualitative results.}
    We show representative restoration results for two datasets:
    ER (upper panels) and
    F-actin (lower panels).
    For each dataset, we show
    \textbf{(a)} the full degraded input with a selected crop (yellow box),
    \textbf{(b)} the degraded crop,
    \textbf{(c)} the corresponding ground truth,
    \textbf{(d--g)} predictions from four competing methods, and
    \textbf{(h)} \BEACON{} ($K=16$).
    For generative posterior models, diagonally split images show the MMSE estimate (upper-left) and a representative posterior sample (lower-right), with MMSE PSNR and sample LPIPS reported, respectively.
    For point-predicting methods, PSNR and LPIPS are reported directly.
}
  \label{fig:qualitative_supp4}
\end{figure*}
}

\newcommand{\figQualitativeSuppFive}{
\begin{figure*}[t]
\centering
  \includegraphics[width=0.8\textwidth]{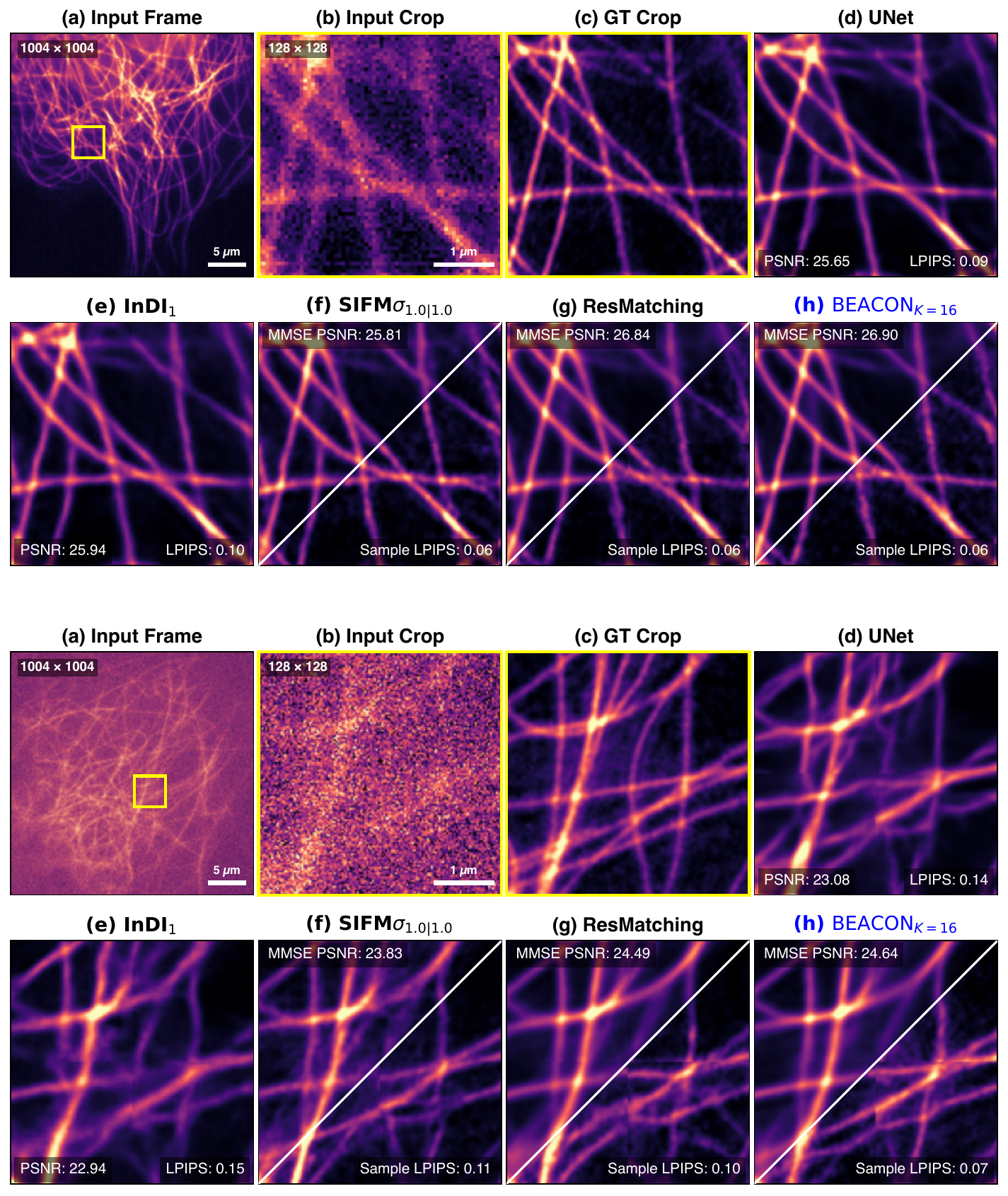}
    \caption{\textbf{Additional qualitative results.}
    We show representative restoration results for two datasets:
    MT (super-resolution) (upper panels) and
    MT Noisy (lower panels).
    For each dataset, we show
    \textbf{(a)} the full degraded input with a selected crop (yellow box),
    \textbf{(b)} the degraded crop,
    \textbf{(c)} the corresponding ground truth,
    \textbf{(d--g)} predictions from four competing methods, and
    \textbf{(h)} \BEACON{} ($K=16$).
    For generative posterior models, diagonally split images show the MMSE estimate (upper-left) and a representative posterior sample (lower-right), with MMSE PSNR and sample LPIPS reported, respectively.
    For point-predicting methods, PSNR and LPIPS are reported directly.
}
  \label{fig:qualitative_supp5}
\end{figure*}
}

%% file: sec/0_abstract.tex
\begin{abstract}
Fluorescence microscopy drives much of modern life science, but accessible imaging setups often yield images degraded by out-of-focus light, limited resolution, and strong signal-dependent noise.
Computational restoration can recover high-quality images from such data, enabling cheaper and faster imaging at lower light dose, provided biologically meaningful structures are preserved.
Guided conditional flow matching has recently shown strong results for microscopy dehazing and computational super-resolution by learning distributions of plausible restorations.
However, it pairs each clean target with an independent standard Gaussian sample, so the starting point of every training path does not leverage the degraded observation.
This arbitrary noise--target coupling can produce conflicting velocity targets and makes the conditional velocity field harder to learn.
We present \BEACON, a \textbf{BE}st-of-$K$ \textbf{A}daptive learned coupling for guided \textbf{CON}ditional flow matching, which learns an observation-conditioned base distribution.
During training, a lightweight conditional noise model proposes $K$ starting points for each degraded input, and the candidate with the lowest flow-matching loss supervises both networks, while a KL term keeps the proposal close to the Gaussian prior.
At inference, a single draw from the learned proposal is transported along the velocity field, requiring neither candidate selection nor the target.
We evaluate \BEACON on ten datasets spanning microscopy dehazing and computational super-resolution against 13 baseline families.
\BEACON achieves the best median perception--distortion rank for every pairing of a fidelity metric (PSNR, MicroMS-SSIM) with a perceptual metric (LPIPS, FID).
Code and pretrained models will be released under a permissive license.
\end{abstract}

%% file: sec/1_intro.tex
\section{Introduction}
\figTeaserMain
Fluorescence microscopy restoration is an ill-posed inverse problem, where a single degraded observation may correspond to multiple plausible high-quality reconstructions. This ambiguity arises from uncertainty in the acquired measurements, which can be constrained, but not necessarily eliminated, by structural priors learned from training data. Deterministic restoration methods, typically optimized using mean squared error (MSE), produce a single prediction that averages over plausible solutions, often resulting in overly smooth reconstructions. As established by the perception--distortion trade-off~\citep{PDT}, optimizing for data fidelity (e.g., high PSNR) can come at the expense of perceptual realism (e.g., low LPIPS or FID), and vice versa. Generative Image Restoration (GIR) methods, such as diffusion models and conditional flow matching, address this ambiguity by learning a conditional distribution over plausible reconstructions, enabling posterior sampling and offering a means to balance data fidelity and perceptual realism~\citep{resmatching2026}.

\figTraining

Recently, Guided Conditional Flow Matching (GCFM), an iterative GIR method, has demonstrated strong performance in microscopy dehazing and computational super-resolution under severe noise conditions~\cite{ray2026hazematching,resmatching2026}, with a favorable position on the perception--distortion frontier.
However, GCFM pairs each target with an independent Gaussian sample: the observation guides the velocity field but not the base--target assignment, which can produce conflicting velocity targets at similar intermediate states~\citep{TrajMin}.

In this work, we introduce \textbf{\BEACON}, which adds a lightweight Conditional Noise Model (CNM) that learns an observation-conditioned noise distribution to construct adaptive base--target couplings (Figure~\ref{fig:overview}).
During training, the CNM proposes $K$ candidate starting points for each degraded input--target pair, and the candidate with the lowest flow-matching loss supervises the velocity model, while KL regularization keeps the learned distribution close to a standard Gaussian prior.
At inference, a single sample from the learned conditional distribution is transported through the velocity field, without best-of-$K$ selection or access to the target image.
We evaluate \BEACON on ten datasets for microscopy dehazing and computational super-resolution, jointly assessing reconstruction fidelity and perceptual realism.

%% file: sec/2_RelatedWork.tex
\section{Related Work}

\miniheadline{Light microscopy image restoration}
Deep Learning-based microscopy restoration uses image data to estimate structures obscured by noise, haze, or limited resolution. Deterministic models~\citep{unet, CARE, pn2v,mimo,mprnet,rcan} produce a single estimate, leaving ambiguity in the restoration unresolved~\citep{floriansReviewWithSulianaEtAl}. Probabilistic approaches model alternative reconstructions, but operate in a single-step prediction mode~\citep{hdn,hvae}. Recent generative iterative approaches like \HazeMatching and \ResMatching apply guided conditional flow matching to dehazing and computational super-resolution, respectively~\cite{ray2026hazematching,resmatching2026}.

\miniheadline{Data-dependent and learned couplings}
In microscopy image restoration, conditioning on the degraded observation is essential to preserve fidelity to the underlying biological structures while generating perceptually realistic predictions~\citep{resmatching2026,ray2026hazematching}. Beyond conditioning the velocity field, the choice of base--target coupling determines the probability paths used to supervise flow matching and influences the learned transport geometry. Prior work has explored optimal-transport-based couplings~\cite{mbOT} and learned forward processes to reduce trajectory intersections and curvature~\cite{learned_conditional,TrajMin}. In particular, Lee et al.~\cite{learned_conditional} learn a Gaussian base distribution conditioned on the target image, but without incorporating the degraded observation. SIFM~\cite{sifm} constructs data-dependent couplings using an explicit degradation operator and a prescribed noise level, which may be unknown or difficult to estimate in microscopy due to signal-dependent noise. Similarly, PMRF~\cite{PMRF} relies on an additional MMSE estimator and a carefully tuned noise parameter.

\miniheadline{Best-of-$K$ and explorative training}
Our best-of-$K$ selection draws inspiration from Implicit Maximum Likelihood Estimation (IMLE)~\cite{IMLE}, which matches training examples to their nearest generated samples to encourage mode coverage, and Explorative Modeling~\cite{ExpModeling}, which optimizes the best among multiple candidates to mitigate mode averaging. Concurrent work on Competitive Flow Matching (CFM)~\cite{peng2026cfm} similarly employs winner-take-all training, but introduces auxiliary latent codes to produce multiple competing velocity predictions at a shared intermediate state, improving the quality--coverage trade-off in image generation. Unlike these approaches, \BEACON performs the competition over source--target couplings rather than over velocity predictions, sampling $K$ candidate starting points from a source distribution conditioned on the degraded observation.

%% file: sec/3_Method.tex
\section{Methods}
\label{sec:method}

\miniheadline{Problem definition}
Following~\cite{ray2026hazematching,resmatching2026}, we consider paired microscopy observations of the same underlying biological sample $s\in\mathcal{S}$, acquired under two imaging conditions $M_0$ and $M_1$.
The degraded input $\mathbf{x}_{M_0}\sim p_{M_0}(\mathbf{x}\mid s)$ and high-quality target $\mathbf{x}_{M_1}\sim p_{M_1}(\mathbf{x}\mid s)$ are drawn from the corresponding conditional image distributions.
Depending on the task, $\mathbf{x}_{M_0}$ represents a hazy or low-resolution microscopy measurement, while $\mathbf{x}_{M_1}$ represents its non-hazy or high-resolution counterpart. Given paired training data $(\mathbf{x}_{M_0},\mathbf{x}_{M_1})\sim p_{\mathrm{data}}$, we seek a conditional generative model of $p(\mathbf{x}_{M_1}\mid\mathbf{x}_{M_0})$. As in~\citep{ray2026hazematching, resmatching2026}, no explicit knowledge of the image degradation operator is required.

\subsection{Guided Conditional Flow Matching}
\label{sec:guided_cfm}

Guided conditional flow matching (GCFM) learns a continuous transport from a standard Gaussian base distribution to the target image distribution while conditioning the velocity field on the degraded measurement~\cite{lipman,sifm}. Let $\mathbf{x}_0\sim p_0=\mathcal{N}(\mathbf{0},\mathbf{I})$ denote a base sample and $t\sim\mathcal{U}[0,1]$ a time.
A network $v_\theta(t,\mathbf{x}_t,\mathbf{x}_{M_0})$ estimates the measurement-guided marginal velocity along the linear interpolant $\mathbf{x}_t$ by minimizing
\begin{equation}
\begin{aligned}
\mathbf{x}_t &= (1-t)\mathbf{x}_0+t\mathbf{x}_{M_1},\\
\mathcal{L}_{\mathrm{GCFM}}(\theta) &= \mathbb{E}\big\|v_\theta(t,\mathbf{x}_t,\mathbf{x}_{M_0})-(\mathbf{x}_{M_1}-\mathbf{x}_0)\big\|^2,
\end{aligned}
\label{eq:gcfm_loss}
\end{equation}
where the expectation is over paired training images, independent $\mathbf{x}_0\sim p_0$, and sampled $t$.
In this standard construction, the base sample is independent of the observed pair, \ie $p(\mathbf{x}_0,\mathbf{x}_{M_0},\mathbf{x}_{M_1})=p_0(\mathbf{x}_0)\,p_{\mathrm{data}}(\mathbf{x}_{M_0},\mathbf{x}_{M_1})$.
Thus, the observation guides the velocity regression but does not influence which noise realization is paired with its target.

\miniheadline{Why the noise--target coupling matters}
Even though each path in Eq.~\eqref{eq:gcfm_loss} is straight, independently sampled base--target pairs can pass through similar intermediate states while prescribing different velocity targets~\citep{TrajMin}. Under squared-error regression, the population-optimal guided velocity is
\begin{equation}
\begin{aligned}
v^\star(t,\mathbf{x}_t,\mathbf{x}_{M_0})
=\mathbb{E}\big[\mathbf{x}_{M_1}-\mathbf{x}_0\,\big|\mathbf{x}_t,\mathbf{x}_{M_0}\big].
\end{aligned}
\label{eq:conditional_marginal_velocity}
\end{equation}
Consequently, nearby or overlapping conditional trajectories with conflicting velocities contribute to an averaged marginal direction. Such intersections have been associated with increased curvature of learned generative trajectories~\cite{TrajMin,mbOT}.

Lee et al.~\cite{TrajMin} further motivate learning the forward process by relating generative trajectory curvature to numerical integration error. This perspective inspires our adaptation of the base--target coupling; our selection criterion is the guided flow-matching loss rather than an explicit curvature penalty.

In microscopy restoration, the observed pair $(\mathbf{x}_{M_0},\mathbf{x}_{M_1})$ is physically meaningful because both images describe the same specimen. In contrast, assigning an independently drawn Gaussian realization to that pair is arbitrary. We therefore retain the experimental observation--target correspondence and adapt only the base--target assignment. Rather than a purely independent coupling, we investigate a measurement-aware coupling.

\subsection{\BEACON: Best-of-$K$ Adaptive Learned Coupling}
\label{sec:beacon}

Figure~\ref{fig:overview} illustrates the \BEACON training framework, including the conditional noise model, best-of-$K$ candidate selection, and alternating updates of the flow and noise models.

\BEACON learns a measurement-conditioned proposal over base samples and explores $K$ candidate couplings per paired training example. Each sampled base induces a candidate interpolation and velocity target; the candidate with the lowest current guided flow-matching error is used for the parameter update. The measured pair $(\mathbf{x}_{M_0},\mathbf{x}_{M_1})$ is never reassigned.

\miniheadline{Conditional noise model}
A conditional noise model $f_\phi$ predicts pixel-wise maps of the mean $\boldsymbol{\mu}_\phi$ and the log-variance $\log\boldsymbol{\sigma}_\phi^2$ that define a diagonal Gaussian proposal, from which $K$ candidates are drawn by reparameterization with independent $\boldsymbol{\epsilon}^{(k)}\sim\mathcal{N}(\mathbf{0},\mathbf{I})$:
\begin{equation}
\begin{aligned}
(\boldsymbol{\mu}_\phi,\boldsymbol{\sigma}_\phi)&=f_\phi(t,\mathbf{x}_{M_0}),\\
q_\phi(\mathbf{x}_0\mid\mathbf{x}_{M_0})&=\mathcal{N}\big(\boldsymbol{\mu}_\phi,\operatorname{diag}(\boldsymbol{\sigma}_\phi^2)\big),\\
\mathbf{x}_0^{(k)}&=\boldsymbol{\mu}_\phi+\boldsymbol{\sigma}_\phi\odot\boldsymbol{\epsilon}^{(k)},\quad k=1,\ldots,K,
\end{aligned}
\label{eq:noise_reparameterization}
\end{equation}
where $\odot$ denotes element-wise multiplication. The noise model shares the architecture of the velocity network, with its time input fixed to $t=0$. Conditioning on $\mathbf{x}_{M_0}$ thus influences both the proposal from which training paths are initiated and the predicted velocity field.

\miniheadline{Candidate transport and best-of-$K$ selection}
Inspired by nearest-sample matching in IMLE~\cite{IMLE} and best-match training in Explorative Modeling~\cite{ExpModeling}, we explore multiple candidate base samples for each fixed observation--target pair and rank their induced paths by guided velocity-regression error.

For each training pair and sampled time $t$, the $k$th candidate defines the interpolant $\mathbf{x}_t^{(k)}=(1-t)\mathbf{x}_0^{(k)}+t\mathbf{x}_{M_1}$, its guided flow-matching loss $\ell^{(k)}$, and the best-of-$K$ (BoK) training objective
\begin{equation}
\begin{aligned}
\ell^{(k)}&=\big\|v_\theta(t,\mathbf{x}_t^{(k)},\mathbf{x}_{M_0})-(\mathbf{x}_{M_1}-\mathbf{x}_0^{(k)})\big\|_2^2,\\
\mathcal{L}_{\mathrm{BoK}}(\theta,\phi)&=\mathbb{E}\big[\min_{1\leq k\leq K}\ell^{(k)}\big],
\end{aligned}
\label{eq:bok_loss}
\end{equation}
where the minimizing index $k^\star$ is selected separately for each observation--target pair and the expectation also includes the random candidate perturbations. For $K=1$, this objective uses a single sample from the learned proposal; fixing that proposal to $\mathcal{N}(\mathbf{0},\mathbf{I})$ recovers standard GCFM. The $min$ is treated as a discrete selection: the gradient flows through the selected candidate and its reparameterized path, not through the selection index itself.

\miniheadline{Prior regularization}
To regularize the conditional proposal toward a standard Gaussian prior, the proposal is updated with
\begin{equation}
\begin{aligned}
\mathcal{L}_{\mathrm{noise}}(\theta,\phi)&=\mathcal{L}_{\mathrm{BoK}}(\theta,\phi)+\beta\,\mathcal{L}_{\mathrm{KL}}(\phi),\\
\mathcal{L}_{\mathrm{KL}}(\phi)&=D_{\mathrm{KL}}\big(q_\phi(\mathbf{x}_0\mid\mathbf{x}_{M_0})\,\|\,\mathcal{N}(\mathbf{0},\mathbf{I})\big),
\end{aligned}
\label{eq:noise_total_loss}
\end{equation}
where $\beta$ controls the balance between candidate-path compatibility and prior regularity.
Conditional KL regularization does not guarantee that the aggregate proposal $q_\phi(\mathbf{x}_0)=\int q_\phi(\mathbf{x}_0\mid\mathbf{x}_{M_0})\,p_{M_0}(\mathbf{x}_{M_0})\,\mathrm{d}\mathbf{x}_{M_0}$ exactly equals $\mathcal{N}(\mathbf{0},\mathbf{I})$.
This is intended: the proposal may deviate from the prior where this lowers the selected flow-matching loss, while the KL term penalizes such deviations, including a collapse of $\boldsymbol{\sigma}_\phi$ towards zero.

\miniheadline{Training and alternating optimization}
Because an untrained velocity model provides unreliable candidate rankings, we first warm up $v_\theta$ using independent Gaussian base samples and the standard GCFM objective in Eq.~\eqref{eq:gcfm_loss}. We then alternate optimization of the velocity and conditional noise networks. For each minibatch of size $B$, the noise network predicts $\boldsymbol{\mu}_\phi$ and $\boldsymbol{\sigma}_\phi$ once per degraded observation. The $K$ candidate noises, interpolants, and velocity targets are constructed in parallel. We fold the candidate dimension into the batch dimension so that all $BK$ candidate paths can be evaluated by a batched velocity-model pass rather than a serial loop over $K$.

The best candidate is selected per original minibatch item.
With the proposal frozen, the velocity update minimizes $\mathcal{L}_{\mathrm{BoK}}$ with respect to $\theta$; with the velocity network frozen, the proposal update minimizes $\mathcal{L}_{\mathrm{noise}}$ in Eq.~\eqref{eq:noise_total_loss} with respect to $\phi$.
The selected candidate is re-evaluated with the counterpart network fixed as needed for each update. Gradient flow through the reparameterization in Eq.~\eqref{eq:noise_reparameterization} enables the proposal to adapt its mean and standard deviation. Non-selected candidates contribute no direct Best-of-$K$ gradient for that selection.
Supp.\ Algorithm~\ref{alg:beacon_training} summarizes the full training procedure.

\miniheadline{Inference and posterior sampling}
At test time, only $\mathbf{x}_{M_0}$ is available.
The conditional noise network is evaluated once per input, and a single $\mathbf{x}_0$ is drawn from the learned proposal as in Eq.~\eqref{eq:noise_reparameterization}.
We then integrate the guided ODE $\mathrm{d}\mathbf{x}_t/\mathrm{d}t=v_\theta(t,\mathbf{x}_t,\mathbf{x}_{M_0})$ from $t=0$ to $t=1$ with $T$ uniform Euler steps, yielding the restored image $\widehat{\mathbf{x}}_{M_1}=\mathbf{x}_{1}$.

Multiple plausible restorations are obtained by independently sampling the initial state and repeating the ODE integration.
The mean $\widehat{\mathbf{x}}_{\mathrm{MMSE}}=\frac{1}{N}\sum_{n=1}^N\widehat{\mathbf{x}}_{M_1}^{(n)}$ of $N$ such predictions approximates the conditional mean of the model's generated distribution, and their pixel-wise variation characterizes predictive variability.

\miniheadline{Computational complexity}
The additional cost of \BEACON is incurred during training, where the velocity model scores $K$ candidate paths per example, so that training-time memory and velocity-model computation grow approximately linearly with $K$.
At inference, no selection is performed, and \BEACON uses the same $T=20$ ODE steps as~\citep{ray2026hazematching,resmatching2026} plus one proposal-network evaluation per input (Supp.\ Section~\ref{supp:compute}).

%% file: sec/4_Experiments.tex
\section{Experiments}

\miniheadline{Datasets}
We evaluate \BEACON on microscopy dehazing (widefield-to-confocal-like predictions) from~\citep{ray2026hazematching} and computational super-resolution under heavy noise from~\citep{resmatching2026} to examine its use across two major restoration tasks in light microscopy. The datasets span biological structures and degradation regimes. We keep the same dataset splits and pre-processing (normalization regime) as mentioned in~\citep{ray2026hazematching} and ~\citep{resmatching2026}, respectively.
For dehazing, we use the five datasets of~\citep{ray2026hazematching}: Zebrafish, Organoids1, Organoids2, Microtubule, and Neuron.
For super-resolution, we use the five datasets of~\citep{resmatching2026}: CCP, ER, F-actin, MT, and MT-Noisy.
MT-Noisy is derived from MT by applying the additional noise scheme introduced in~\citep{resmatching2026}.
Note that Microtubule (dehazing) and MT (super-resolution) are distinct datasets.

\miniheadline{Baselines}
We compare \BEACON with deterministic restoration networks and variational, diffusion-based, and flow-based generative methods. \HazeMatching and \ResMatching~\cite{ray2026hazematching,resmatching2026} provide the direct guided flow-matching comparisons for assessing the effect of adapting the base--target assignment. For all other baselines, we follow the implementation details mentioned in~\citep{ray2026hazematching} and ~\citep{resmatching2026}, respectively. Additional details related to baselines are given in Supp.\ Section~\ref{supp:baselines}.

\ranking
\miniheadline{Evaluation metrics}
\label{sec:evaluation_metrics}
Restoration quality requires both fidelity and perceptual evaluation. We measure distortion using PSNR and MicroMS-SSIM, and perceptual quality using LPIPS and FID. To examine their trade-offs, we report PSNR--LPIPS, PSNR--FID, MicroMS-SSIM--LPIPS, and MicroMS-SSIM--FID pairs.

For a relative summary, we independently normalize distortion and perceptual scores to $[0,1]$, with lower values indicating better performance. The empirical ideal point $(0,0)$ combines the best observed score on each axis; it need not correspond to an attainable reconstruction. For method $m$ with normalized scores $d_m$ and $p_m$, the combined summary score adds the distance $a_m$ to this point and the distance $b_m$ to the diagonal,
\begin{equation}
    r_m = a_m+b_m,\quad a_m=\sqrt{d_m^2+p_m^2},\quad b_m=\frac{|d_m-p_m|}{\sqrt{2}}.
    \label{eq:pd_rank_score}
\end{equation}
This score combines proximity to the empirical ideal with a penalty for imbalance between normalized errors (Figure~\ref{fig:nipd}). Because the penalty can increase when one metric improves, the score does not preserve all Pareto improvements and must be interpreted alongside the original metric pairs.

We rank methods by ascending $r_m$ for each dataset and metric pair, then summarize ranks across the available test datasets separately for each metric pair. Missing dataset results are not imputed. These rankings depend on the compared methods and normalization; they summarize relative balance rather than absolute restoration quality.

\miniheadline{Uncertainty calibration}
\label{sec:calibration}
\BEACON enables stochastic posterior sampling by drawing different initial noise realizations from its learned observation-conditioned base distribution. To assess whether the resulting predictive uncertainty reflects the true reconstruction error, we evaluate the calibration of its posterior samples against high-quality ground-truth images. Following~\cite{ray2026hazematching,Microsplit}, we estimate pixel-wise uncertainty from the variability across multiple predictions and group pixels into bins according to their predicted uncertainty. For each bin, we compute the root mean variance (RMV) and root mean squared error (RMSE), representing predicted uncertainty and observed reconstruction error, respectively. We then linearly calibrate RMV against RMSE and assess their agreement using reliability diagrams. A well-calibrated model yields values close to the identity line ($y=x$), indicating that higher predictive uncertainty corresponds to larger reconstruction errors. Further details related to calibration are provided in Supp.\ Section~\ref{supp:calibration}.

\miniheadline{Implementation details}
\BEACON uses a \UNet~\citep{diffusinOpenAI} velocity network with 3.70M parameters and a smaller variant of the same architecture as the conditional noise network, with 234K parameters. Both networks are implemented in PyTorch~\citep{pytorch} and trained for 200 epochs on NVIDIA V100 GPUs using the Adam optimizer with alternating updates. Before alternating training, we warm up the velocity network for one epoch using independent samples from the standard Gaussian base distribution $\mathcal{N}(\mathbf{0},\mathbf{I})$. We use training patches of size $64\times64$ or $128\times128$ and set the KL regularization weight to $\beta=0.4$. The $K$ candidates are folded into the batch dimension for velocity evaluation. Inference samples one starting point from the proposal and integrates the guided ODE with $T=20$ steps. We use the \texttt{torchCFM}~\citep{torchCFM} library for interpolant computation and ODE integration.
We use the same training, validation, and test sample counts as \HazeMatching~\citep{ray2026hazematching} for dehazing and \ResMatching~\citep{resmatching2026} for super-resolution. Full training details are provided in Supp.\ Section~\ref{supp:training}.

\miniheadline{Evaluation} Predictions on full-sized images use inner tiling~\cite{Microsplit}, with the tile size matching the training patch size and a 50\% overlap.

%% file: sec/5_Results.tex
\section{Results}

\subsection{Quantitative Results}

\figRankMain

\figQualitative

We compare \BEACON with the baselines across datasets and metric pairs to assess the effect of adaptive base--target assignment. Figure~\ref{fig:rank_main} summarizes the main quantitative comparison for $K=16$, using dataset-wise perception--distortion ranks for all four metric pairs introduced in Section~\ref{sec:evaluation_metrics}. Both \BEACON and the task-specific \ResMatching/\HazeMatching baselines are evaluated on ten datasets, where distortion metrics are computed on the MMSE ($N=50$) and perceptual metrics are computed for each individual prediction and then averaged over the number of samples as done in~\citep{ray2026hazematching,resmatching2026}. Their comparison assesses the combined effect of the learned conditional proposal and best-of-$K$ selection relative to independent Gaussian initialization.
Per-dataset scores for all methods are reported in Supp.\ Section~\ref{supp:results}.
Compared with its direct guided flow-matching counterpart, \BEACON improves PSNR on six of the ten datasets, is within $0.02$\,dB on three, and is lower on CCP ($-0.48$\,dB).
FID improves on seven datasets, most strongly on Zebrafish ($2.17\rightarrow0.62$) and Organoids1 ($0.91\rightarrow0.24$), and is worse on Neuron, ER, and F-actin.
LPIPS is lower on four datasets and higher on six, with the largest increase on ER ($0.222\rightarrow0.273$).
The gains are most consistent for dehazing, where PSNR and FID improve on four of the five datasets, whereas results for super-resolution are mixed.

In the aggregate ranking (Figure~\ref{fig:rank_main}), \BEACON attains the lowest median rank for all four metric pairs, while its mean rank is slightly worse than that of the guided flow-matching baseline for MicroMS-SSIM--FID.

The ranking therefore supports improved relative perception--distortion balance under the chosen score, rather than superiority on every individual metric. Rank distributions must also be interpreted in light of dataset coverage: some competing methods have results on only one or five datasets, whereas \BEACON and the direct guided baseline each have ten. As discussed in Section~\ref{sec:evaluation_metrics}, the ranking depends on normalization and the compared methods and does not establish improvement on both original metric axes.

\subsection{Qualitative Results}

Figure~\ref{fig:qualitative} shows representative crops for two dehazing datasets (Organoids1, Zebrafish) and two super-resolution datasets (MT, ER), comparing \BEACON with the four best-ranked competing methods on each dataset.
On these crops, the MMSE estimate of \BEACON reaches the highest PSNR among the shown methods on Organoids1, Zebrafish, MT, and ER, and its posterior sample the lowest LPIPS on MT and ER and the tied-lowest on Zebrafish.
Visually, \BEACON recovers distinct punctate signals and thin filamentous extensions in Organoids1 while suppressing diffuse background fluorescence, resolves individual microtubule filaments and their crossings in MT, and reconstructs the interconnected tubular ER network including narrow branches and junctions.

Although we employ inner tiling~\citep{Microsplit} to mitigate boundary artifacts, some tiling artifacts remain visible in the reconstructed images. Future work could explore adapting approaches such as SWITi~\citep{carrara2026switi} to further reduce these artifacts and improve spatial consistency across tile boundaries.

\subsection{Ablations and Analysis}

We analyze the effect of the number of candidates $K$ and whether best-of-$K$ selection alone explains the gains of \BEACON.

\miniheadline{Choice of $K$}
We evaluate different $K$ values across all 10 validation datasets (Supp.\ Fig.~\ref{fig:supp_k_ablation}). 
$K=16$ achieves both the highest average PSNR (30.84\,dB) and lowest average FID (0.4373). 
We therefore use \BEACON with $K=16$ for all main results, although differences between values of $K$ are small.

\miniheadline{Number of candidates $K$ on test data}
Increasing $K$ expands candidate exploration but also increases training cost.
We compare $K=1$ with $K>1$ under the learned proposal on the test data to assess the contribution of best-of-$K$ selection, using $K\in\{1,2,4,8,12,16,20\}$ (Supp.\ Fig.~\ref{fig:ablation_k}), with ranks computed against the full set of compared methods.
$K=16$ attains the lowest or tied-lowest median rank for PSNR--FID, PSNR--LPIPS, and MicroMS-SSIM--FID, whereas $K=1$ and $K=2$ reach lower medians for MicroMS-SSIM--LPIPS. 
$K=20$ has the highest median rank in all four pairs, and the ordering of intermediate values is not monotonic in $K$.
With a single training run per setting, differences between neighboring values of $K$ should not be over-interpreted; notably, $K=1$, which uses the learned proposal without candidate selection, remains competitive throughout.

\miniheadline{Learned base distribution}
Supp.\ Fig.~\ref{fig:conditional_noise_visualization} visualizes the learned proposal ($K=16$) on three representative Zebrafish patches. The predicted variance maps exhibit spatial structure correlated with the underlying morphology, with lower variance at prominent cellular structures, while the predicted mean remains close to zero. 

\miniheadline{Effect of best-of-$K$ on baseline methods}
To assess whether Best-of-$K$ training alone explains the improvements of \BEACON, we incorporate the same candidate selection strategy ($K=16$) into \HazeMatching and \ResMatching while retaining their standard Gaussian initialization. Although this improves the baselines in some metric pairs, \BEACON achieves lower median ranks than the Best-of-$K$ baselines in two comparisons (PSNR--LPIPS and PSNR--FID) and ties their median ranks in the other two (MicroMS-SSIM--LPIPS and MicroMS-SSIM--FID), although its mean ranks are slightly higher on the latter. This suggests that learning an observation-conditioned base distribution provides additional benefits beyond Best-of-$K$ selection for PSNR-based perception--distortion trade-offs, while improvements are not uniform across metrics.
A detailed comparison is provided in Supp.\ Fig.~\ref{fig:ablation_best_k}.

\miniheadline{Inference-step ablation}
Similar to \HazeMatching and \ResMatching~\citep{ray2026hazematching,resmatching2026}, \BEACON allows users to adjust the number of inference steps $T$ without retraining, thereby varying the Euler step size and the resulting perception--distortion trade-off. Users can choose fewer or more steps according to their preference for perceptual quality or reconstruction fidelity, with the corresponding operating points assessed using LPIPS and PSNR, respectively. Supp.\ Section~\ref{supp:ablation_steps} considers this trade-off on two datasets, comparing \BEACON ($K=16$) with the corresponding \HazeMatching and \ResMatching baselines.

\subsection{Calibration Analysis}

\figCalibMain

Predictive variation does not by itself establish calibrated uncertainty. Figure~\ref{fig:calib_main} evaluates \BEACON with $K=16$ on two dehazing datasets (Organoids1 and Organoids2) and two super-resolution datasets (MT and ER), using the binned RMV--RMSE protocol in Section~\ref{sec:calibration}. We compare the native uncertainty estimates with those obtained after linear calibration, with the identity line indicating agreement between predicted variability and observed reconstruction error.

For both organoid datasets, the unscaled curves increasingly lie above the identity line as uncertainty grows, indicating that native predictive variability underestimates reconstruction error. Linear calibration brings these curves closer to the identity over much of the range, although deviations remain in the highest-uncertainty bins. The super-resolution datasets exhibit a different pattern. On ER, the unscaled curve lies above the identity at intermediate RMV and below it at high RMV; the scaled curve retains substantial deviations, so a global linear correction does not fully capture this relationship. On MT, the unscaled curve follows the identity relatively closely at low and intermediate RMV but falls below it at higher RMV. Scaling shifts the high-uncertainty bins toward and then above the identity, while introducing overestimated uncertainty in lower bins. The relation between predictive variability and reconstruction error is thus dataset dependent.

%% file: sec/6_Discussion.tex
\section{Discussion and Limitations}

\BEACON replaces the observation-independent base--target assignment of GCFM with observation-conditioned starting points that are selected by their current flow-matching loss.
This may ease velocity regression, but a lower selected loss alone does not establish reduced trajectory curvature or improved restoration.
We therefore evaluate fidelity and perceptual quality jointly, since perceptual realism is insufficient for microscopy when reconstructed details disagree with the measured structure.
Likewise, predictive variability describes variation in the learned model and requires separate validation, such as the calibration analysis above, before it can be interpreted as calibrated uncertainty.

Best-of-$K$ selection increases training computation and introduces the hyperparameters $K$ and $\beta$.
It also uses the target during training, whereas inference draws directly from the conditional proposal, so the selected training samples can differ in distribution from inference samples.
All reported results are therefore obtained with the target-free inference procedure.

\section{Conclusion}

We introduced \BEACON, which combines an observation-conditioned Gaussian proposal with best-of-$K$ training-path selection for guided flow matching, adapting the base--target assignment while preserving measured microscopy input--target pairs.
On microscopy dehazing and computational super-resolution, \BEACON achieves the best median perception--distortion rank across ten datasets, with the most consistent gains on dehazing.

%% file: sec/9_llm_statement.tex
\section*{LLM Usage Statement}
\label{sec:llm}
We used LLM-based assistants (Claude, Anthropic; GPT models, OpenAI) for coding, writing, and typesetting.
\emph{Code and experiments:} writing and debugging evaluation and plotting scripts, and cross-checking reported numbers against experiment outputs.
\emph{Figures:} \LaTeX{} layout of figures and floats; the microscopy images and predictions shown are unedited data and outputs of the evaluated models, and no figure content is AI-generated.
\emph{Main text and supplement:} drafting and editing of text, including language, structure, and \LaTeX{} formatting, condensing the text to the page limit, consistency of notation and references, and literature search.
\emph{Ideation:} no research ideas; only suggestions for control experiments and for presenting results clearly within limited space.
All methods, experiments, and results were designed, carried out, and verified by the authors, who take full responsibility for the content.

%% file: sec/8_Appendix.tex
\newpage
\onecolumn 
\begin{center}
    {\Large \textbf{\BEACON: Learning Observation-Conditioned Starting Points for Guided Conditional Flow Matching in Microscopy}}
\end{center}

\vspace{0.5em}

\begin{center}
    {\Large \textbf{\textit{Supplementary Material}}}
\end{center}

\setcounter{table}{0} 
\renewcommand{\thetable}{S\arabic{table}} 

\renewcommand{\theequation}{s.\arabic{equation}}
\setcounter{equation}{0}

\appendix
\setcounter{section}{0}
\renewcommand{\thesection}{S\arabic{section}} 
\renewcommand{\theHsection}{S\arabic{section}} 
\counterwithin{figure}{section} 
\renewcommand{\theHfigure}{\theHsection.\arabic{figure}} 
\input{supple_sec/00_training}

\input{supple_sec/0_compute}

\input{supple_sec/1_calibration}
\input{supple_sec/2_baselines}
\input{supple_sec/4_Ablation}
\input{supple_sec/6_dataset}
\input{supple_sec/3_results}

%% file: supple_sec/00_training.tex
\section{Training Details}
\label{supp:training}

\subsection{Training Algorithm}

\begin{algorithm}[h]
\caption{\textsc{BEACON} Training}
\label{alg:beacon_training}
\begin{algorithmic}[1]
\Require $v_\theta$, $f_\phi$, candidate count $K$, KL weight $\beta$
\State Warm up $v_\theta$ using Eq.~\eqref{eq:gcfm_loss}
\Repeat
\State Draw minibatch $(\mathbf{x}_{M_0},\mathbf{x}_{M_1})$
\State Draw $t\sim\mathcal{U}[0,1]$ per pair
\State $(\boldsymbol{\mu}_\phi,\boldsymbol{\sigma}_\phi)\gets f_\phi(t=0,\mathbf{x}_{M_0})$
\State Draw $K$ independent $\boldsymbol{\epsilon}^{(k)}\sim\mathcal{N}(0,I)$ per pair
\State Construct all $\mathbf{x}_0^{(k)}$ with Eq.~\eqref{eq:noise_reparameterization}
\State Construct all interpolants $\mathbf{x}_t^{(k)}$
\State Fold candidate axis into batch axis
\State Evaluate $\ell^{(1)},\ldots,\ell^{(K)}$ in parallel
\State Select $k^\star\gets\arg\min_k\ell^{(k)}$ per pair
\State Freeze $\phi$; update $\theta$ using selected loss
\State Freeze $\theta$; update $\phi$ using selected loss $+\beta\mathcal{L}_{\mathrm{KL}}$
\Until{convergence}
\State \Return $v_\theta,f_\phi$
\end{algorithmic}
\end{algorithm}

\subsection{Training settings}

We trained \BEACON separately on each dataset subset for 200 epochs with a batch size of 16, using separate Adam optimizers for the velocity and conditional noise networks with a fixed learning rate of $10^{-4}$. Training used $128 \times 128$ images ($64 \times 64$ for the neuron subset). After one warm-up epoch using standard Gaussian noise, we enabled conditional noise learning with a fixed $K=16$ candidates per sample and minimum-loss candidate selection. The KL regularization weight $\beta$ was set to $0.4$. Training times were sampled uniformly from 20 equally spaced points in $[0,1]$, and the checkpoint with the lowest validation loss after warm-up was selected for evaluation. The inference setting is same as in~\citep{ray2026hazematching,resmatching2026}, where the first $t=0$ draw is drawn from the conditional noise model.

%% file: supple_sec/0_compute.tex
\section{Compute Details}
\label{supp:compute}

The conditional proposal network is evaluated once per degraded image, whereas the velocity model scores $K$ candidate paths per training example.
Folding the candidate dimension into the batch axis lets these $K$ evaluations run in parallel, but increases training-time memory consumption and total velocity-model computation approximately linearly with $K$.
Best-of-$K$ selection is not performed at inference, which uses the same number of ODE steps and velocity-network evaluations as~\citep{ray2026hazematching,resmatching2026} ($T=20$), plus one proposal-network evaluation per input.

\newpage
\subsection{Training time per $K$}
\figTrainTimeK

\subsection{Inference runtime comparison of \BEACON with baselines}

We compare the inference efficiency of \BEACON against
\HazeMatching and other restoration baselines in
Table~\ref{tab:runtime_comparison}. Despite introducing
an additional conditional noise model, \BEACON incurs
negligible computational overhead during inference.
The noise model contains only approximately 234K
parameters and is evaluated once per image patch to
generate the initial noise sample. The subsequent
restoration is performed by the velocity model using
the same number of integration steps as \HazeMatching.
Consequently, \BEACON achieves a comparable inference
runtime (4.23\,s versus 4.75\,s for \HazeMatching),
demonstrating that the benefits of learned noise
initialization come at virtually no additional
inference cost. The reported numbers for all the baselines are from~\cite{ray2026hazematching}.

\begin{table}[h]
\centering
\small
\begin{tabular}{lccc}
\toprule
\textbf{Model} &
\textbf{Mean (5) [sec.]} &
\textbf{Per Forward Pass [sec.]} &
\textbf{Params (M)} \\
\midrule
RL            & 13.6       & N/A     & N/A \\
Elements      & 3.233      & N/A     & N/A \\
Lightning     & 1.0 (est.) & N/A     & N/A \\
UNet          & 0.2944     & N/A     & 13.39 \\
MIMO-UNet     & 0.6358     & N/A     & 6.80 \\
MPRNet        & 5.4713     & N/A     & 20.12 \\
RCAN          & 8.6271     & N/A     & 15.29 \\
ESRGAN        & 1.1839     & N/A     & 16.70 \\
Restormer     & 4.5067     & N/A     & 26.12 \\
InDI$_1$      & 7.1888     & N/A     & 2.44 \\
InDI$_{20}$   & 143.3886   & 7.16943 & 2.44 \\
\HVAE         & 1.0118     & N/A     & 5.67 \\
SIFM          & 4.3120     & 0.2156  & 3.70 \\
\HazeMatching & 4.7498     & 0.23749 & 3.70 \\
\textbf{\BEACON}
              & 4.2337\textsuperscript{*}
              & 0.21168
              & 3.94\textsuperscript{**} \\
\bottomrule
\end{tabular}

\vspace{1mm}
\begin{minipage}{\linewidth}
\footnotesize
\textsuperscript{*} The marginal runtime difference
is negligible and may arise from measurement variability.

\textsuperscript{**} The velocity-prediction model has
3.70M parameters. The reported 3.94M corresponds
to the noise-prediction mode, whose noise network
is evaluated only once per patch.
\end{minipage}

\caption{\textbf{Inference runtime comparison on Organoids1 Dataset.}
Mean inference time over five runs, per-forward-pass runtime,
and number of model parameters.}
\label{tab:runtime_comparison}
\end{table}

%% file: supple_sec/1_calibration.tex
\section{Calibration}
\label{supp:calibration}

\subsection{Calibration Method}

We evaluate uncertainty calibration following the procedure
described in \cite{ray2026hazematching, resmatching2026}.
For each test image, we generate $k=50$ posterior samples
and compute the pixel-wise standard deviation $\sigma(p)$
and the mean prediction $\hat{y}(p)$.
Pixels are sorted by their predicted uncertainty and
partitioned into $l=50$ equally populated bins.
For each bin $B_j$, we compute the root mean variance
(RMV) and root mean squared error (RMSE) as
\begin{equation}
\begin{aligned}
\mathrm{RMV}(j) &=
\sqrt{\frac{1}{|B_j|}
\sum_{p\in B_j}\sigma^2(p)},\\
\mathrm{RMSE}(j) &=
\sqrt{\frac{1}{|B_j|}
\sum_{p\in B_j}(\hat{y}(p)-y(p))^2},
\end{aligned}
\end{equation}
where $y$ denotes the ground truth.
Using the validation set, we fit a linear calibration
mapping by minimizing
\begin{equation}
\min_{\alpha>0,\,\beta_{calib}}
\left\|\mathrm{RMSE}-
(\alpha\,\mathrm{RMV}+\beta_{calib})\right\|_2^2.
\end{equation}
The learned mapping is applied to test-set
uncertainties without modifying the predictions.
All calibration experiments are performed on
normalized data.

\subsection{Sample efficiency for calibration}
\figCalibEfficiency

\newpage
\subsection{Calibration Results}
\figCalibrationSupp

%% file: supple_sec/2_baselines.tex
\section{Baselines}
\label{supp:baselines}

We follow the baseline configurations of \HazeMatching~\citep{ray2026hazematching} for dehazing and
\ResMatching~\citep{resmatching2026} for computational super-resolution.
Both comparisons include single-output predictors and generative models
that can sample multiple restorations per input. The dehazing comparison
also includes classical and commercial deconvolution methods reported in~\citep{ray2026hazematching}.

%% file: supple_sec/4_Ablation.tex
\clearpage
\section{Ablations}
\label{supp:ablation}

\subsection{Effect of $K$}

\figAblationK
\clearpage

\subsection{Effect of Best-of-$K$ on Baseline Methods}
\figAblationBestK
\figNoiseanalysis
\clearpage
\figValKSelection

\newpage
\subsection{Effect of the Number of Inference Steps}
\label{supp:ablation_steps}
\figStepsPSNRLPIPS
The number of inference steps $T$ provides a user-adjustable control over the perception--distortion operating point without changing the trained model. With uniform Euler integration over $[0,1]$, reducing $T$ increases the step size $\Delta t=1/T$ and reduces inference cost, whereas increasing $T$ yields finer integration at greater computational cost. As in \HazeMatching and \ResMatching~\citep{ray2026hazematching,resmatching2026}, this allows users to explore settings that favor perceptual quality or reconstruction fidelity rather than relying solely on the default $T=20$.

The comparison, shown in Figure~\ref{fig:steps_psnr_lpips}, considers two datasets, one for dehazing (Organoids1) and one for super-resolution (ER), pairing \BEACON ($K=16$) with \HazeMatching and \ResMatching, respectively. PSNR and LPIPS characterize the distortion and perceptual quality at each value of $T$. The preferred setting depends on the dataset and the desired balance between these criteria; increasing $T$ need not improve both metrics simultaneously.

\subsection{Effect of Posterior Sample Averaging}
\label{sec:posterior_averaging}

We investigate how the number of posterior samples
affects reconstruction fidelity by computing the PSNR
of their pixel-wise mean. As shown in
Fig.~\ref{fig:samples_psnr}, increasing the number of
averaged samples improves PSNR for both \BEACON and
the corresponding \HazeMatching and \ResMatching baselines,
with diminishing returns beyond approximately 20--30
samples. Interestingly, although \BEACON exhibits lower
single-sample PSNR, it surpasses both baselines when
averaging as few as five predictions and consistently
maintains higher fidelity thereafter. This suggests
that \BEACON's individual sample predictions exhibit errors
that are more effectively reduced through averaging,
resulting in a more accurate posterior-mean estimate.

\figSamplesPSNR

%% file: supple_sec/6_dataset.tex
\clearpage
\section{Dataset}
\label{supp:dataset}

We evaluate on five dehazing datasets from \HazeMatching and five \ResMatching
subsets for computational super-resolution. The latter comprise the four
subsets evaluated in \ResMatching and the microtubule subset without added
noise (MT), available through the official \ResMatching repository.
Table~\ref{tab:dataset_details} summarizes the image splits and training
patches.

\begin{table*}[h]
    \centering
    \small
    \begin{tabular*}{0.94\textwidth}{@{\extracolsep{\fill}}lrrrlrl@{}}
        \toprule
        & \multicolumn{4}{c}{\textbf{Image pairs}}
        & \multicolumn{2}{c}{\textbf{Training patches}} \\
        \cmidrule(lr){2-5}\cmidrule(lr){6-7}
        \textbf{Dataset} & \textbf{Train} & \textbf{Val.} & \textbf{Test}
        & \textbf{Train size} & \textbf{Count} & \textbf{Size} \\
        \midrule
        \multicolumn{7}{@{}l}{\textbf{Dehazing}} \\
        Zebrafish   & 15 & 3 & 3 & $1024 \times 1024$ & 3{,}000 & $128 \times 128$ \\
        Organoids1  & 15 & 2 & 3 & $1024 \times 1024$ & 3{,}000 & $128 \times 128$ \\
        Organoids2  & 15 & 2 & 3 & $1024 \times 1024$ & 3{,}000 & $128 \times 128$ \\
        Microtubule\textsuperscript{a}
                    & 52 & 5 & 5 & $624 \times 624$ & 3{,}016 & $128 \times 128$ \\
        Neuron      & 5{,}800 & 100 & 100 & $64 \times 64$ & 5{,}800 & $64 \times 64$ \\
        \midrule
        \multicolumn{7}{@{}l}{\textbf{Computational super-resolution (BioSR)}} \\
        CCP         & 39 & 5 & 10 & $1004 \times 1004$ & 3{,}120 & $128 \times 128$ \\
        ER          & 53 & 5 & 10 & $1004 \times 1004$ & 4{,}240 & $128 \times 128$ \\
        F-actin     & 35 & 5 & 10 & $1004 \times 1004$ & 2{,}800 & $128 \times 128$ \\
        MT-Noisy    & 40 & 5 & 10 & $1004 \times 1004$ & 3{,}200 & $128 \times 128$ \\
        MT          & 40 & 5 & 10 & $1004 \times 1004$ & 3{,}200 & $128 \times 128$ \\
        \bottomrule
    \end{tabular*}

    \smallskip
    \parbox{0.94\textwidth}{\footnotesize
        \textsuperscript{a}\,The Microtubule dehazing validation and test
        images are $512 \times 512$ pixels.}

    \caption{Dataset splits and training patch counts for dehazing and
    computational super-resolution. Train, Val., and Test count paired
    images before patch extraction. Sizes are in pixels.}
    \label{tab:dataset_details}
\end{table*}

The Microtubule dehazing validation and test images are
$512 \times 512$ pixels. For Neuron, each $64 \times 64$ training image
is used directly as a patch.

%% file: supple_sec/3_results.tex
\clearpage
\section{Results}
\label{supp:results}

\subsection{Perception--Distortion Plots}

\figQuantitativeOne
\figQuantitativeTwo
\figQuantitativeThree
\figQuantitativeFour
\figQuantitativeFive

\clearpage

\section{Qualitative Results}
\label{supp:qual_results}

\figQualitativeSuppOne
\figQualitativeSuppTwo
\figQualitativeSuppThree
\figQualitativeSuppFour
\figQualitativeSuppFive